\pdfoutput=1
\documentclass[dvipsnames, 11pt]{article}
\usepackage{graphicx} 
\RequirePackage[top=2.25cm, bottom=2.5cm, left=2.5cm, right=2.5cm, columnsep=0.65cm]{geometry}
\usepackage{multirow}
\usepackage{amsmath,amssymb,bm}
\usepackage{booktabs}
\usepackage{xcolor}
\usepackage[numbers]{natbib}
\usepackage{wrapfig}
\usepackage{hyperref}
\usepackage{subcaption}
\usepackage{caption}
\usepackage{makecell}
\renewcommand{\thefootnote}{\fnsymbol{footnote}}
\def\method{\textrm{Vidi2}} 
\def\methodnew{\textrm{Vidi2.5}} 
\def\vidiedit{\textrm{Vidi-Edit}} 
\def\ie{\textit{i.e.}}
\def\eg{\textit{e.g.}}

\usepackage{xcolor}

\definecolor{MyDarkRed}{rgb}{0.8,0.02,0.02}
\definecolor{MyDarkBlue}{rgb}{0.02,0.02,0.8}
\definecolor{MyDarkGreen}{rgb}{0.1,0.8,0.1}
\definecolor{darkgreen}{rgb}{0.0, 0.5, 0.0}
\definecolor{ElectricPurple}{rgb}{0.5, 0.0, 1.0}

\newcommand{\revise}[1]{#1}

\newif\ifdraft
\drafttrue      

\title{\includegraphics[width=.8cm]{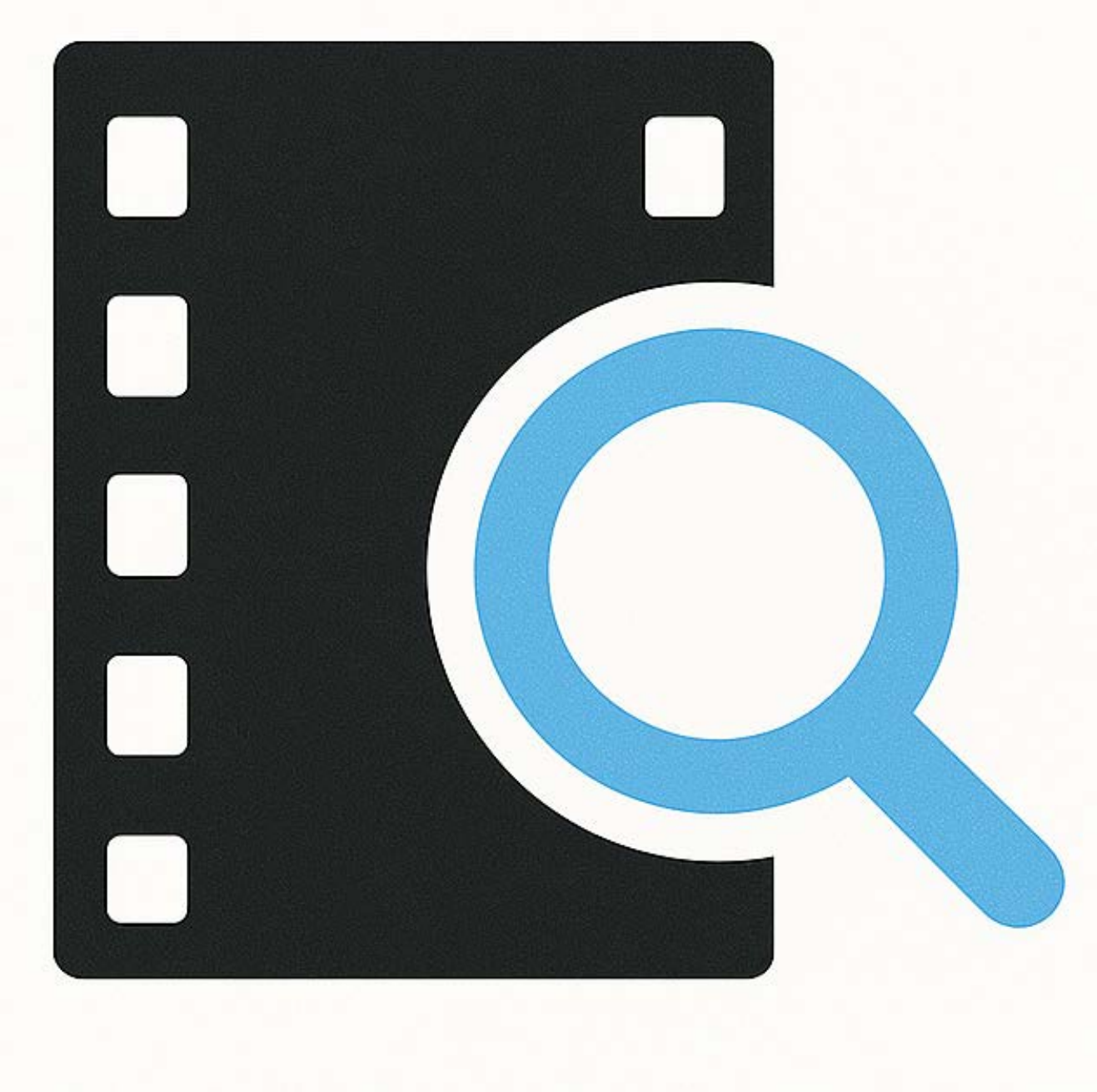} \huge\bfseries \methodnew{}: Large Multimodal Models for Video Understanding and Creation}
\author{Intelligent Editing Team\footnote{A detailed contributor list can be found in Section~\ref{sec:contributor}.} , Intelligent Creation, ByteDance Inc. \\ San Jose/Seattle, US \\ \url{https://bytedance.github.io/vidi-website/}}
\date{}

\begin{document}

\maketitle

\vspace{-0.89cm}
\section*{Abstract}
\label{sec:abstract}
\vspace{-0.15cm}
\small{
Video has emerged as the primary medium for communication and creativity on the Internet, driving strong demand for scalable, high-quality video production.
Vidi models continue to evolve toward next-generation video creation and have achieved state-of-the-art performance in multimodal temporal retrieval (\textbf{TR}).
In its second release, \method{} advances video understanding with fine-grained spatio-temporal grounding (\textbf{STG}) and extends its capability to video question answering (\textbf{Video QA}), enabling comprehensive multimodal reasoning.
Given a text query, \method{} can identify not only the corresponding timestamps but also the bounding boxes of target objects within the output time ranges.
This end-to-end spatio-temporal grounding capability enables potential applications in complex editing scenarios, such as plot understanding, automatic view switching, etc.
To enable comprehensive evaluation of STG in practical settings, we introduce a new benchmark, \textbf{VUE-STG}, which offers critical improvements over existing STG datasets in video duration, query format, annotation quality, and evaluation metric.
In addition, we upgrade the previous VUE-TR benchmark to \textbf{VUE-TR-V2}, achieving a more balanced duration and query distribution.
Remarkably, the \method{} model substantially outperforms leading proprietary systems, such as Gemini 3 Pro Preview and GPT-5, on both VUE-TR-V2 and VUE-STG, while achieving competitive results with popular open-source models with similar scale on video QA benchmarks. The latest \textbf{\methodnew{}} offers significantly stronger STG capability and slightly better TR and Video QA performance over \method{} by RL training. This update also introduces a \textbf{\methodnew{}-Think} model to handle plot understanding with complex plot reasoning. To comprehensively evaluate the performance of plot understanding, we propose a new benchmark named \textbf{VUE-PLOT} with two tracks, \textit{Character} and \textit{Reasoning}. Notably, \textbf{\methodnew{}-Think} outperforms Gemini 3 Pro Preview on fine-grained character understanding with comparable performance on complex plot reasoning. 
Furthermore, we demonstrate the effectiveness of \methodnew{} on a challenging real-world application, video editing planning. The post-trained variant \textbf{\vidiedit{}} can generate structured editing plans specifying narrative structure, audio attributes, and visual editing intent, illustrating its strong potential for complex real-world tasks.
}

\begin{figure*}[!hbtp]
    \centering
    \vspace{-0.25cm}
    \includegraphics[trim={30 10 0 0},clip,width=\linewidth]{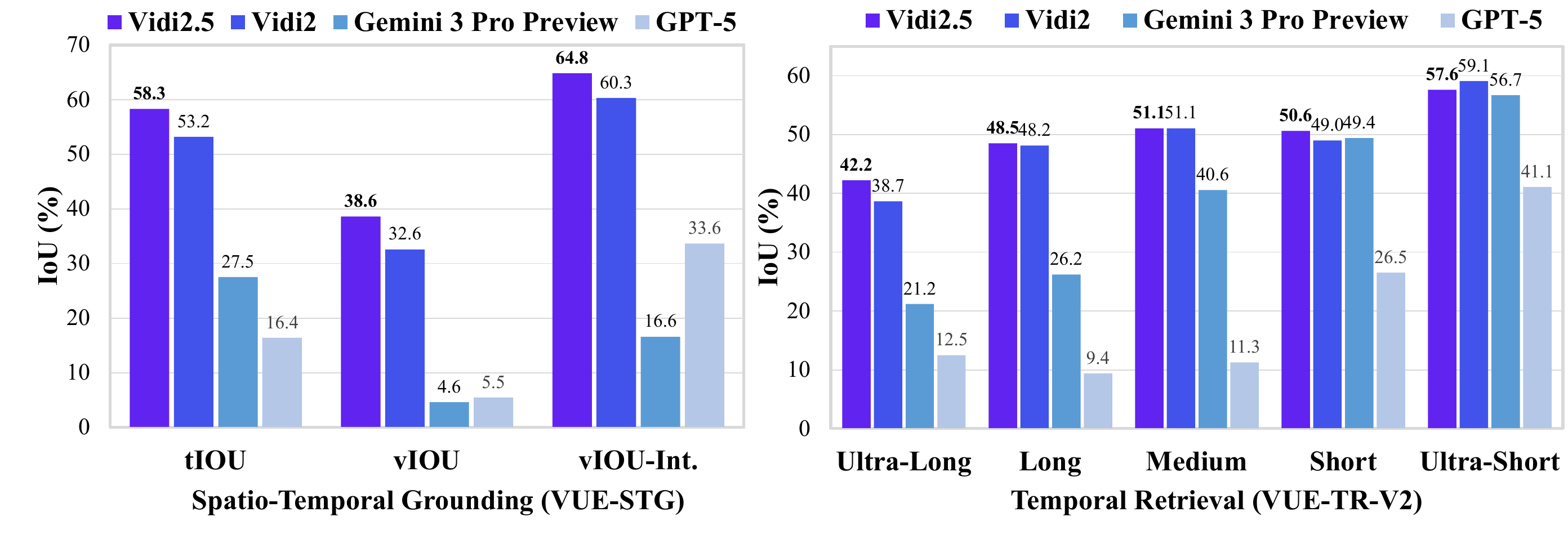}
    \vspace{-0.7cm}
    \caption{Spatio-temporal grounding and temporal retrieval results on the proposed benchmarks.}
    \vspace{-0.5cm}
    \label{fig:teaser}
\end{figure*}

\newpage
\section{Introduction}
\label{sec:introduction}
As an inherently multimodal medium, video has become a dominant channel for online communication and information sharing.
However, producing high-quality videos remains challenging for most users, particularly when performing trimming or editing operations on mobile devices.
The first release of {\bf Vidi}~\cite{team2025vidi} demonstrated strong temporal understanding across text, visual, and audio modalities, revealing its potential for automatic video trimming and understanding.
To advance toward next-generation video creation systems capable of handling complex editing scenarios, Vidi models have been continually evolving to support more comprehensive multimodal perception and reasoning capabilities.

\begin{figure*}[!htbp]
    \centering
    \vspace{-0.2cm}
    \includegraphics[width=\linewidth]{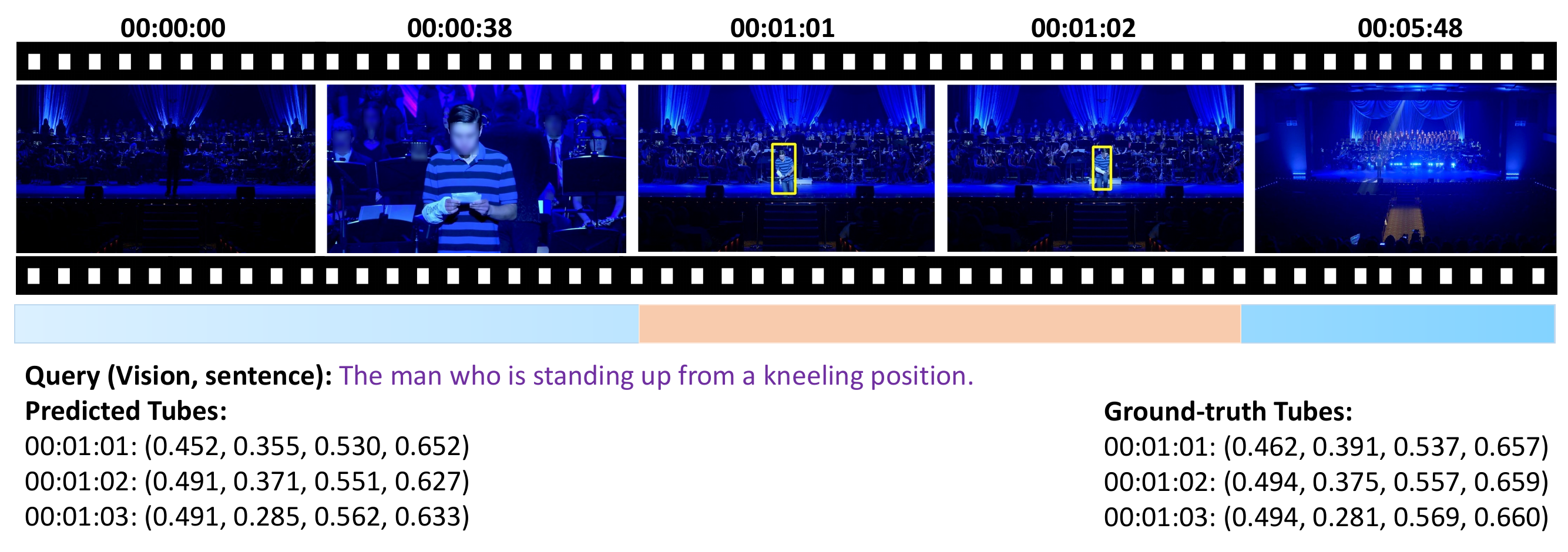}
    \caption{Examples of spatio-temporal grounding queries and their corresponding time ranges and object tubes (timestamps with bounding boxes, shown in yellow). Bounding boxes are expressed in percentage coordinates. The example video has a total duration of $387$ seconds (\ie, $\texttt{06:27}$), and the query has been converted into a noun-style format. Facial regions are blurred to protect privacy.}
    \label{fig:stvg_intro}
\end{figure*}

The second release, {\bf \method{}}~\cite{team2025vidi2}, introduces for the first time an end-to-end spatio-temporal grounding (STG) capability, identifying not only the temporal segments corresponding to a text query but also the spatial locations of relevant objects within those frames.
Notably, existing academic models~\cite{gu2024context,gu2025knowing,jin2022embracing,lin2023collaborative,su2021stvgbert,wasim2024videogrounding} and state-of-the-art industrial systems such as Gemini 3 Pro Preview~\cite{Gemini,comanici2025gemini} and GPT-5~\cite{OpenAI2025GPT5SystemCard} are not yet capable of producing such fine-grained grounding results in a unified text output format (see Fig.~\ref{fig:teaser}).
As illustrated in Fig.~\ref{fig:stvg_intro}, the query ``a man standing up from a kneeling position'' represents a particularly difficult case involving multiple people in a dark scene.
\method{} accurately localizes the corresponding time ranges and distinguishes the target person from others through precise bounding boxes in an end-to-end manner. It requires a comprehensive understanding of the visual and temporal dynamics of the described action.
This fine-grained spatio-temporal perception highlights \method{}'s potential to enable advanced editing workflows, such as plot-level understanding, automatic multi-view switching, and composition-aware reframing for professional video creation.

To enable comprehensive evaluation of spatio-temporal grounding (STG) in realistic scenarios, we introduce a new benchmark, VUE-STG, featuring videos that range from $10$ seconds to $30$ minutes, substantially longer than existing STG datasets~\cite{chen2019weakly,tang2021human,yamaguchi2017spatio,zhang2020does}.
Compared to prior work, VUE-STG includes videos and annotated tubes of much greater temporal extent, along with more complex, multimodal queries that often incorporate audio cues to assist temporal localization.
All timestamps and bounding boxes are manually annotated with high precision, ensuring reliable spatio-temporal alignment.
In addition, we upgrade the previous VUE-TR benchmark to VUE-TR-V2, featuring more human-authored, free-form queries and a more balanced video-length distribution with an increased proportion of long and ultra-long videos, making the task more realistic and challenging.

On the VUE-STG benchmark, \method{} surpasses leading proprietary models (Gemini 3 Pro Preview and GPT-5) by a substantial margin, demonstrating state-of-the-art spatio-temporal grounding capability among large multimodal models.
Compared with the previous version, \method{} also shows significant improvements in temporal retrieval, particularly for long videos.
On the updated VUE-TR-V2 benchmark, it outperforms Gemini 3 Pro Preview\footnote{Note that Gemini 3 Pro shows a notable improvement over Gemini-2.5-Pro-0325 on temporal retrieval.} and GPT-5 by a wide margin across ``Medium'' to ``Ultra-Long'' video categories, confirming its leading performance in temporal reasoning.
Furthermore, \method{} extends its capability to generic video question answering, marking a step toward a foundation model for comprehensive video understanding with strong spatio-temporal perception. 
\begin{figure*}
    \centering
    \includegraphics[width=\linewidth]{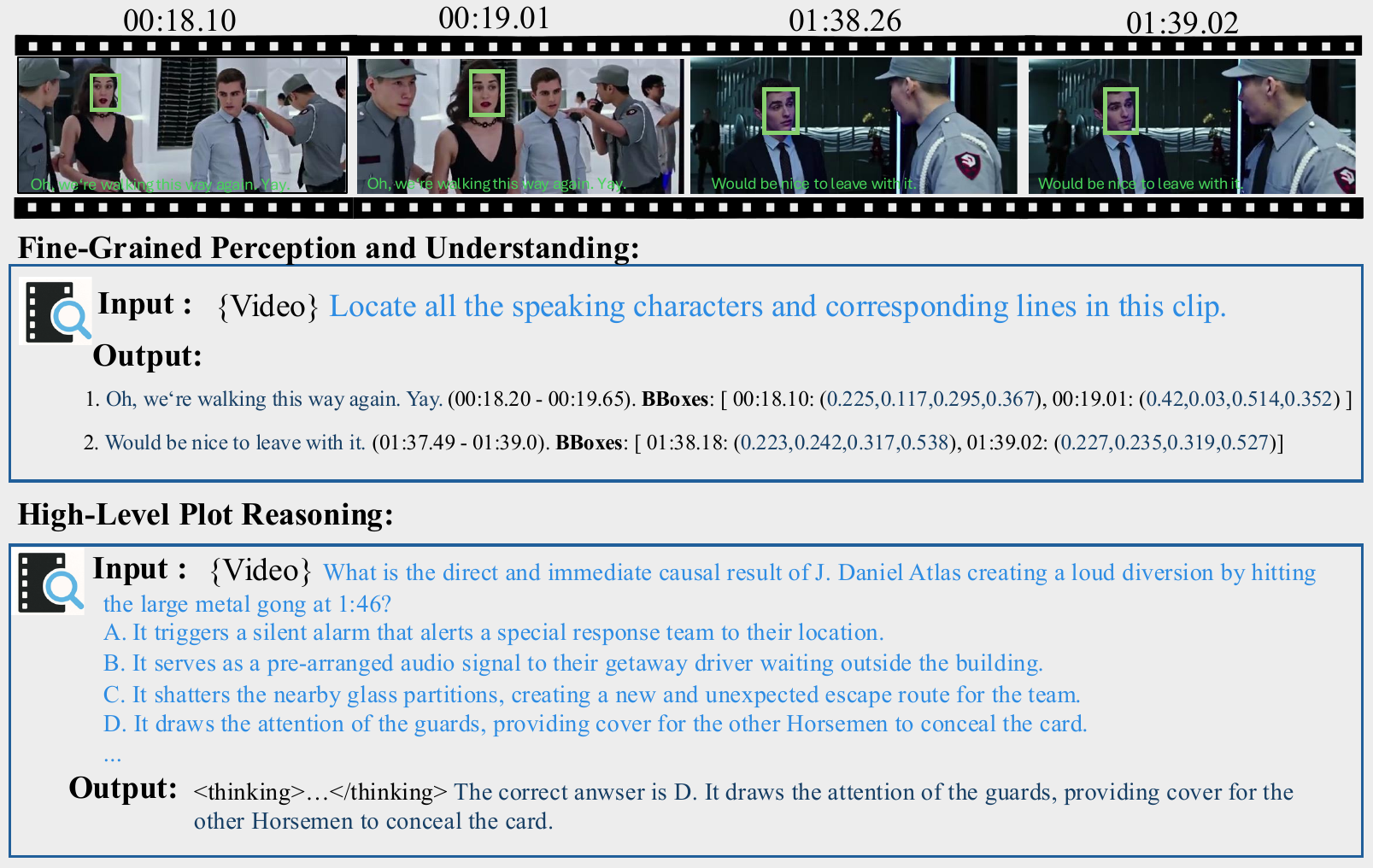}
    \caption{Example of plot understanding and reasoning tasks in VUE-PLOT. The \textit{Character} track requires dense speaker localization and speech recognition, while the \textit{Reasoning} track evaluates narrative understanding, character relationships, and editing techniques via multiple-choice visual question answering. Speakers' faces and spoken lines are shown on the video for better visualization.}
    \label{fig:application_plot}
\end{figure*}

\revise{\textit{The latest \methodnew{} introduces several major updates in Section \ref{sec:introduction}, \ref{sec:vue_plot}, \ref{sec:exp_plot}, \ref{sec:applications} to introduce new models, benchmarks and applications. Complementary results and description are added in Section \ref{sec:overview}, \ref{sec:experiment}.} The base \methodnew{} model introduces multimodal Reinforcement Learning (RL) training with verifiable rewards and consistently boosts STG, TR, and Video QA performance. To further enhance multimodal reasoning capabilities for complex scenarios, \eg, plot understanding and reasoning, we introduce a thinking model for the \methodnew{} model family. Inspired by DeepSeek-R1~\cite{guo2025deepseek} and OpenAI o1~\cite{jaech2024openai}, \methodnew{}-Think supports inference-time scaling to produce higher-quality responses. Unlike these models, \methodnew{}-Think is optimized using a combination of verifiable rewards and generative reward prediction, specifically designed to capture fine-grained details in videos and enable complex plot reasoning.}

\revise{
To comprehensively evaluate models on both fine-grained perception and high-level plot understanding, we introduce a new benchmark: VUE-PLOT. VUE-PLOT is formed by two tracks, \textit{Character} and \textit{Reasoning}. \textit{Character} focuses on real-world videos and assesses fine-grained character perception by requiring models to densely localize the speaking individuals and recognize their spoken content. \textit{Reasoning} targets high-level plot understanding and reasoning, challenging models to reason about narrative dynamics, character relationships, and editing techniques through multiple-choice visual question answering on informative video clips. Notably, our \methodnew{}-Think model outperforms Gemini 3 Pro on character grounding and speech recognition, while achieving competitive performance on complex plot reasoning tasks.
}

In addition to model and benchmark advances, \methodnew{} is also applied to practical video creation workflows. We study video editing planning as a representative application, implemented via a post-trained variant, \vidiedit{}, which transforms raw video assets and optional user intents into structured editing plans. These plans explicitly specify narrative structure, voiceover content, audio attributes, and visual editing intent, serving as high-level specifications for downstream video creation. This application illustrates how \methodnew{} can be used to support complex, real-world video editing workflows beyond standard video understanding settings.

\section{Model Overview}
\label{sec:overview}
\method{} builds on the foundation of Vidi~\cite{team2025vidi} with substantial upgrades in architecture, training data, and task capabilities, most notably the support for spatio-temporal grounding (STG) and video question answering (Video QA). These improvements collectively enable fine-grained multimodal understanding across text, visual, and audio modalities.

\paragraph{Architecture:} 
\method{} retains the multimodal architecture of Vidi~\cite{team2025vidi}, designed to jointly process text, visual, and audio inputs, while introducing key enhancements in both the encoder and LLM backbone (\eg, Gemma-3~\cite{team2025gemma}).
All results presented in this report are based on the 12B-parameter configuration.
To achieve robust performance across videos of different lengths, we re-design the adaptive token compression strategy, improving the balance between short and long video representation efficiency.
Additionally, a single image is treated as a $1$-second silent video, ensuring a unified encoding interface for both image and video inputs.

\paragraph{Training Data:} 
\method{} follows a training pipeline similar to the previous Vidi version but scales up in both data volume and data diversity.
While synthetic data remains essential for coverage and stability, increasing the proportion of real video data significantly enhances performance across all video-related tasks.
The supervised fine-tuning (SFT) dataset for temporal retrieval has been expanded and refined in both quality and quantity.
Furthermore, we incorporate additional generic QA data during SFT to strengthen open-ended reasoning on images and videos, improving the model's versatility on downstream applications such as highlight detection, dense captioning, and video summarization. \revise{Additional human annotated data is collected for RL training of \methodnew{} on verifiable tasks.}

\paragraph{Spatio-Temporal Grounding:} 
To support the newly introduced spatio-temporal grounding (STG) capability, we introduce task-specific data across all training stages.
Specifically, we leverage existing image-level spatial grounding datasets to synthesize large-scale spatio-temporal video grounding pairs, effectively bridging spatial and temporal alignment.
In addition, we curate a substantial collection of real-world video STG annotations, which contribute significantly to the model's ability to localize both time ranges and bounding boxes with high precision.

\paragraph{Video Question Answering:}
The previous Vidi~\cite{team2025vidi} model demonstrated the effectiveness of the temporal-aware multimodal alignment training stage (see Section 4 in~\cite{team2025vidi}) for temporal retrieval.
\method{} further validates the generalizability of this multimodal alignment strategy to generic video QA.
By incorporating video QA data during the post-training phase, \method{} is able to answer both visual and auditory questions with high accuracy.
We also observe that scaling up multimodal alignment data consistently improves overall video QA performance.

\paragraph{Thinking Model:} 
\revise{
\methodnew{} introduces a new thinking model, \methodnew{}-Think, to enhance plot reasoning in videos via inference-time scaling. \methodnew{}-Think is trained using a highly customized reinforcement learning framework built on VERL \cite{sheng2024hybridflow}. The training data consist of diverse multimodal instruction-following examples covering video perception, audio understanding, narrative comprehension, and professional filming and editing techniques. These instructions include both multiple-choice visual question answering tasks and open-ended reasoning tasks. The model is optimized using a combination of verifiable rewards and large language models acting as judges to evaluate the quality of predicted rationales, generated responses, and their mutual consistency.
}
\section{Evaluation Benchmark}
\label{sec:evaluation}
To comprehensively evaluate the performance and generalization of \method{}, we consider three representative categories of video understanding tasks: Spatio-Temporal Grounding (STG), Temporal Retrieval (TR), and Video Question Answering (Video QA).
The first two are evaluated on our newly proposed benchmarks (\ie, VUE-STG and VUE-TR-V2), which are designed to assess fine-grained temporal and spatio-temporal reasoning under realistic editing and understanding scenarios.
For video QA, we adopt widely used public benchmarks, including LVBench~\cite{wang2025lvbench}, LongVideoBench~\cite{wu2024longvideobench}, and VideoMME~\cite{fu2025video}.
All of these video QA datasets employ a multiple-choice question format, which enables standardized, objective, and reproducible evaluation of multimodal reasoning performance.
Together, these datasets provide a unified and comprehensive evaluation protocol that spans retrieval, grounding, and video reasoning. \revise{We further introduce VUE-PLOT along with Vidi2.5 for complex plot understanding in real-world video editing scenarios.}

\subsection{VUE-STG}
We introduce \textbf{VUE-STG}, a new \textbf{V}ideo \textbf{U}nderstanding \textbf{E}valuation benchmark specifically designed to advance \textbf{S}patio-\textbf{T}emporal \textbf{G}rounding in real-world scenarios. VUE-STG addresses four critical aspects often overlooked in prior academic benchmarks \cite{chen2019weakly,tang2021human,yamaguchi2017spatio,zhang2020does}: video duration, query format, annotation quality, and metric design. To ensure the accurate performance signal, all ground-truth labels are manually curated, which yields significantly more accurate and consistent labels than existing benchmarks.

\begin{table}[!htbp]
    \centering
    \begin{tabular}{c| c c c |c}
    \toprule[1.5pt]
    \multirow{2}{*}{Duration Category} & Ultra-short & Short & Medium &  \multirow{2}{*}{Total}\\
     & $<1$ min & $1-10$ mins & $10-30$ mins  \\
     \hline
    \# Videos & 126 & 294 & 562 & 982 \\
    \# Queries (Tubes) & 206 & 461 & 933 & 1,600 \\
    \# Boxes & 1,063 & 3,741 & 7,343 & 12,147 \\
    Video Hours  & 0.82 & 26.28 & 177.69  & 204.79 \\
    \bottomrule[1.5pt]
    \end{tabular}
    \caption{Video duration distribution of the proposed VUE-STG benchmark. The benchmark covers a wide range of video lengths, from ultra-short ($<1$ min) clips to medium ($10-30$ mins) videos, enabling comprehensive evaluation of models across diverse real-world scenarios.}
    \label{tab:VUE-STG}
\end{table}

\subsubsection{Data Distribution} 
Following VUE-TR \cite{team2025vidi}, VUE-STG is built using publicly available videos, and the annotations are made accessible for open research. As presented in Table \ref{tab:VUE-STG}, the benchmark consists of $1,600$ queries across $982$ videos, spanning over $204.79$ hours. It supports attribute-based slicing for fine-grained performance analysis as follows.

\begin{itemize}
    \item \textbf{Video Duration.} Unlike existing datasets limited to short videos, the video length varies from $10$ seconds to $30$ minutes, covering most of the duration ranges encountered in real-world scenarios. To enable duration-wise evaluation, we categorize videos into three balanced buckets following the definition of VUE-TR \cite{team2025vidi}: ultra-short ($<1$ min), short ($1-10$ mins), and medium ($10-30$ mins). 

    \item \textbf{Query.} Each query in VUE-STG is carefully verified by humans, which clearly indicates the target object described in the video, aiming to minimize potential ambiguity in grounding. When a descriptive sentence naturally involves multiple objects, we adjust the phrasing to better highlight the intended target. For example, instead of using an ambiguous expression such as ``a player is loaded into an ambulance,'' we adopt more explicit formulations such as ``the ambulance which the player is being loaded into'' or ``the player who is loaded into the ambulance,'' depending on which object is referenced. This annotation strategy substantially reduces semantic ambiguity and improves the reliability of query-object associations. 

    \item \textbf{Tube.} Each target object is annotated as a spatio-temporal tube at $1$ frame per second, ensuring consistent localization quality across long-form videos. Object-size statistics are computed based on the average bounding box area of each tube, resulting in a balanced distribution, as shown in Table~\ref{tab:vue_stvg_object_size}. Furthermore, Table~\ref{tab:vue_stvg_gt_duration} summarizes the distribution of the temporal duration of the tubes. While most tubes span $3$ to $10$ seconds, both short transient events and long-duration activities are included to support a comprehensive evaluation. We note that not all tubes in VUE-STG are strictly temporally contiguous. In real-world scenarios, target objects may disappear due to camera cuts, occlusions, or transitions between scenes. To better reflect these conditions, we preserve such temporally fragmented tubes rather than enforcing artificial continuity. These fragmented cases allow models to be evaluated under more realistic long-form video dynamics, where objects may intermittently leave and re-enter the visible field.
\end{itemize}

\begin{table}[t]
\centering
\begin{tabular}{c| c c c }
\toprule
Object Size &
\begin{tabular}[c]{@{}c@{}}Small\\$<10\%$\end{tabular} &
\begin{tabular}[c]{@{}c@{}}Medium\\$10\%-30\%$\end{tabular} &
\begin{tabular}[c]{@{}c@{}}Large\\$>30\%$\end{tabular} \\
\midrule
\# Queries (Tubes)      & 746 & 511 & 343 \\
\bottomrule
\end{tabular}
\caption{Object-size distribution of annotated tubes in the VUE-STG benchmark. 
Object size is defined as the average bounding-box area within each tube, normalized by the video frame area.}
\label{tab:vue_stvg_object_size}
\end{table}

\begin{table}[t]
\centering
\begin{tabular}{c| c c c }
\toprule
Tube Duration &
\begin{tabular}[c]{@{}c@{}}Micro-Short\\$<3s$\end{tabular} &
\begin{tabular}[c]{@{}c@{}}Ultra-Short\\$3s\!-\!10s$\end{tabular} &
\begin{tabular}[c]{@{}c@{}}Short\\$10s\!-\!60s$\end{tabular} \\
\midrule
\# Queries (Tubes)      & 179 & 1013 & 408 \\
\bottomrule
\end{tabular}
\caption{Temporal duration distribution of annotated tubes in the VUE-STG benchmark. 
Tubes cover short to long temporal extents, from brief moments ($<3$s) 
to longer activities ($10$--$60$s), enabling fine-grained analysis of temporal grounding performance.}
\label{tab:vue_stvg_gt_duration}
\end{table}

\subsubsection{Evaluation Metrics}
\label{sssec:stvg_metrics}
We evaluate both \emph{temporal grounding accuracy} and
\emph{spatio-temporal grounding accuracy} to assess model performance.
The temporal metrics measure how well the predicted time spans align with the ground-truth temporal annotations, whereas the spatio-temporal metrics evaluate the localization quality of predicted bounding boxes across time.
All results are reported as sample-wise averages on the VUE-STG test set.
Among these, $\mathrm{vIoU}$ serves as the primary ranking metric, jointly capturing both temporal and spatial alignment accuracy.

\medskip
\noindent\textbf{Tube representation.}
A spatio-temporal tube is modeled as a time-dependent bounding box, \ie,
\begin{equation}
    B(t) = \Bigl(x_0(t),\, y_0(t),\, x_1(t),\, y_1(t)\Bigr),
\end{equation}
where $(x_0(t), y_0(t))$ and $(x_1(t), y_1(t))$ denote the top-left and
bottom-right coordinates. Let $B^{\mathrm{pred}}(t)$ and $B^{\mathrm{gt}}(t)$ denote the predicted and
ground-truth tubes, defined on temporal supports
$T_{\mathrm{pred}}, T_{\mathrm{gt}} \subset \mathbb{R}$.
In practice, the temporal axis is discretized by uniform sampling, \ie, we use the sampling rate $1$ frame per second throughout this work.

\medskip
\noindent\textbf{Bounding-box IoU (\texorpdfstring{$\mathrm{bIoU}$}{bIoU}).}
We define
\begin{equation}
    \mathrm{bIoU}(B_1, B_2)
    = \frac{\mathrm{Area}(B_1 \cap B_2)}
           {\mathrm{Area}(B_1 \cup B_2)}
\end{equation} 
as the spatial overlap metric between two bounding boxes $B_1$ and $B_2$.

\medskip
\medskip
\paragraph{Temporal metrics:}
Temporal grounding metrics evaluate how well the predicted time interval aligns with the ground-truth annotated temporal extent, including \texorpdfstring{$\mathrm{tIoU}$}{tIoU}, \texorpdfstring{$\mathrm{tP}$}{tP} and \texorpdfstring{$\mathrm{tR}$}{tR}. Let
$T_{\cap} = T_{\mathrm{pred}} \cap T_{\mathrm{gt}}$ and
$T_{\cup} = T_{\mathrm{pred}} \cup T_{\mathrm{gt}}$.
Temporal precision and recall (\texorpdfstring{$\mathrm{tP}$}{tP} and \texorpdfstring{$\mathrm{tR}$}{tR}) measures how much of the predicted time span overlaps with the ground truth, while temporal recall measures how much of the ground truth duration is covered by the prediction:
\[
    \mathrm{tP}
    =
    \frac{\lvert T_{\cap} \rvert}{\lvert T_{\mathrm{pred}} \rvert},
    \qquad
    \mathrm{tR}
    =
    \frac{\lvert T_{\cap} \rvert}{\lvert T_{\mathrm{gt}} \rvert}.
\]
Meanwhile, temporal IoU (\texorpdfstring{$\mathrm{tIoU}$}{tIoU}) measures the alignment between the predicted and ground-truth
time intervals using an IoU-style ratio:
\[
    \mathrm{tIoU}
    =
    \frac{\lvert T_{\cap} \rvert}
         {\lvert T_{\cup} \rvert}.
\]

\medskip
\paragraph{Spatio-temporal metrics:}
Spatio-temporal metrics evaluate how well the model localizes the target object throughout the video, including \texorpdfstring{$\mathrm{vIoU}$}{vIoU}, \texorpdfstring{$\mathrm{vIoU\text{-}Int.}$}{vIoU-Int}, \texorpdfstring{$\mathrm{vP}$}{vP}, and \texorpdfstring{$\mathrm{vR}$}{vR}.

{\noindent \textbf{Frame-level IoU.}}
For any $t \in T_{\mathrm{pred}} \cup T_{\mathrm{gt}}$, we define the
frame-level IoU as
\[
    \mathrm{IoU}_t
    =
    \begin{cases}
        \mathrm{bIoU}\!\left(B^{\mathrm{pred}}(t),\, B^{\mathrm{gt}}(t)\right),
            & t \in T_{\cap}, \\[6pt]
        0,  & \text{otherwise},
    \end{cases}
\]
where $T_{\cap} = T_{\mathrm{pred}} \cap T_{\mathrm{gt}}$ is the temporal
intersection.

{\noindent \textbf{Accumulated IoU over the temporal intersection.}}
Since $\mathrm{IoU}_t = 0$ outside $T_{\cap}$, the accumulated IoU over any
time domain reduces to the sum over $T_{\cap}$. We define
\[
    S = \sum_{t \in T_{\cap}} \mathrm{IoU}_t .
\]

{\noindent \textbf{Spatio-temporal precision and recall
(\texorpdfstring{$\mathrm{vP}$}{vP} and \texorpdfstring{$\mathrm{vR}$}{vR}).}
We measure the average frame-level IoU over the predicted and ground-truth
temporal spans:
\[
    \mathrm{vP}
    = \frac{S}{\lvert T_{\mathrm{pred}} \rvert},
    \qquad
    \mathrm{vR}
    = \frac{S}{\lvert T_{\mathrm{gt}} \rvert}.
\]

{\noindent \textbf{Spatial IoU over the temporal union (\texorpdfstring{$\mathrm{vIoU}$}{vIoU}).}
To evaluate spatio-temporal accuracy over the entire time span where either
tube exists, we average the frame-level IoU over the temporal union
$T_{\cup} = T_{\mathrm{pred}} \cup T_{\mathrm{gt}}$:
\[
    \mathrm{vIoU}
    = \frac{S}{\lvert T_{\cup} \rvert}.
\]

{\noindent \textbf{Spatial IoU over the temporal intersection
(\texorpdfstring{$\mathrm{vIoU\text{-}Int.}$}{vIoU-Int}).}
For completeness, we also compute the average IoU over the time interval
where both tubes are defined as
\[
    \mathrm{vIoU\text{-}Int.}
    = \frac{S}{\lvert T_{\cap} \rvert}.
\]

\subsection{VUE-TR-V2}
Following the settings of VUE-TR \cite{team2025vidi}, we introduce an updated version, VUE-TR-V2, to provide a more balanced video length distribution and better query quality of the evaluation dataset.

\paragraph{Data Distribution:} 
As shown in Table~\ref{tab:VUE-TR}, VUE-TR-V2 contains a comparable number of videos to VUE-TR, but the total duration increases substantially from $107.87$ hours to $311.11$ hours.
The modality and format distributions (Figure~\ref{fig:distribution}) remain consistent with the original benchmark.
To better reflect real-world conditions, VUE-TR-V2 introduces a greater proportion of long and ultra-long videos, resulting in a more balanced length distribution where short and medium clips no longer dominate the dataset. It covers diverse practical scenarios such as movie mashups and long-form content.
\begin{table}[t]
    \centering
    \resizebox{\linewidth}{!}{
    \begin{tabular}{c| c c c c c |c}
    \toprule[1.5pt]
    \multirow{2}{*}{Duration Category} & Ultra-short & Short & Medium & Long & Ultra-long & \multirow{2}{*}{Total}\\
     & $<1$ min & $1-10$ mins & $10-30$ mins & $30-60$ mins & $>60$ mins \\
     \hline
    \# Videos & 66 & 230 &  315 & 197 & 39 & 847 \\
    \# Queries & 134 & 390 & 430 & 426 & 220 & 1,600 \\
    Video Hours  & 0.74 & 19.85 & 97.08 & 145.95 & 47.50 & 310.72 \\
    \bottomrule[1.5pt]
    \end{tabular}}
    \caption{Duration distribution of videos in the proposed VUE-TR-V2 evaluation benchmark. The dataset covers a wide range of video lengths, from ultra-short clips ($<1$ minute) to ultra-long videos ($>1$ hour), enabling comprehensive evaluation of temporal retrieval models across diverse real-world scenarios. The total video length ($310.72$ hours) of VUE-TR-V2 is increased to over $2.8$ times of the video length ($107.87$ hours) in the previous VUE-TR benchmark.}
    \label{tab:VUE-TR}
\end{table}

\begin{figure}[!hbtp]
    \centering
    \vspace{-0.2cm}
    \includegraphics[width=.6\linewidth]{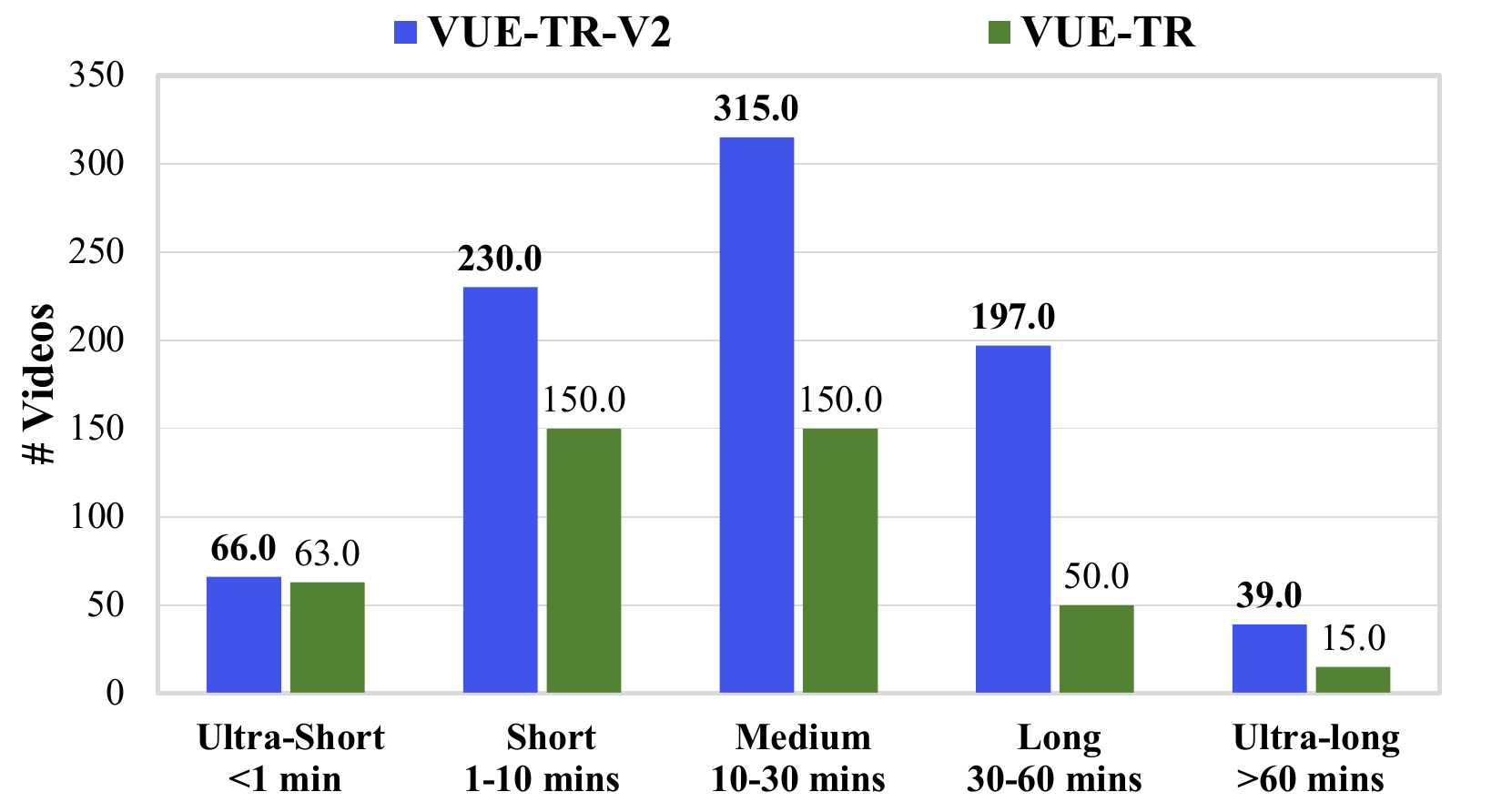}
    \caption{The video distribution comparison between the VUE-TR-V2 and VUE-TR benchmarks. }
    \label{fig:length_distribution}
\end{figure}

\begin{figure}[!htbp]
    \centering
    \vspace{-0.4cm}
    \includegraphics[width=0.48\linewidth]{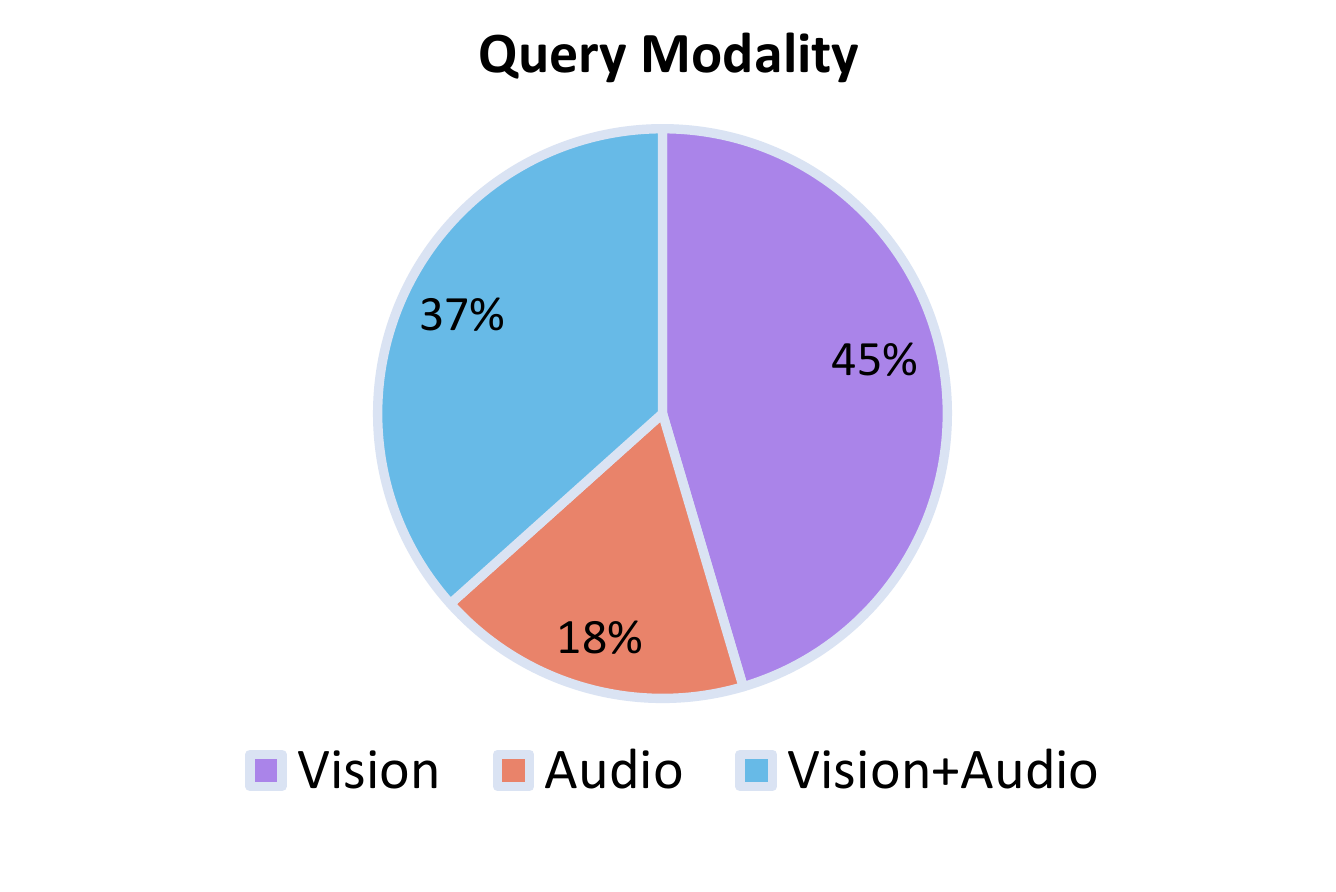}
    \includegraphics[width=0.48\linewidth]{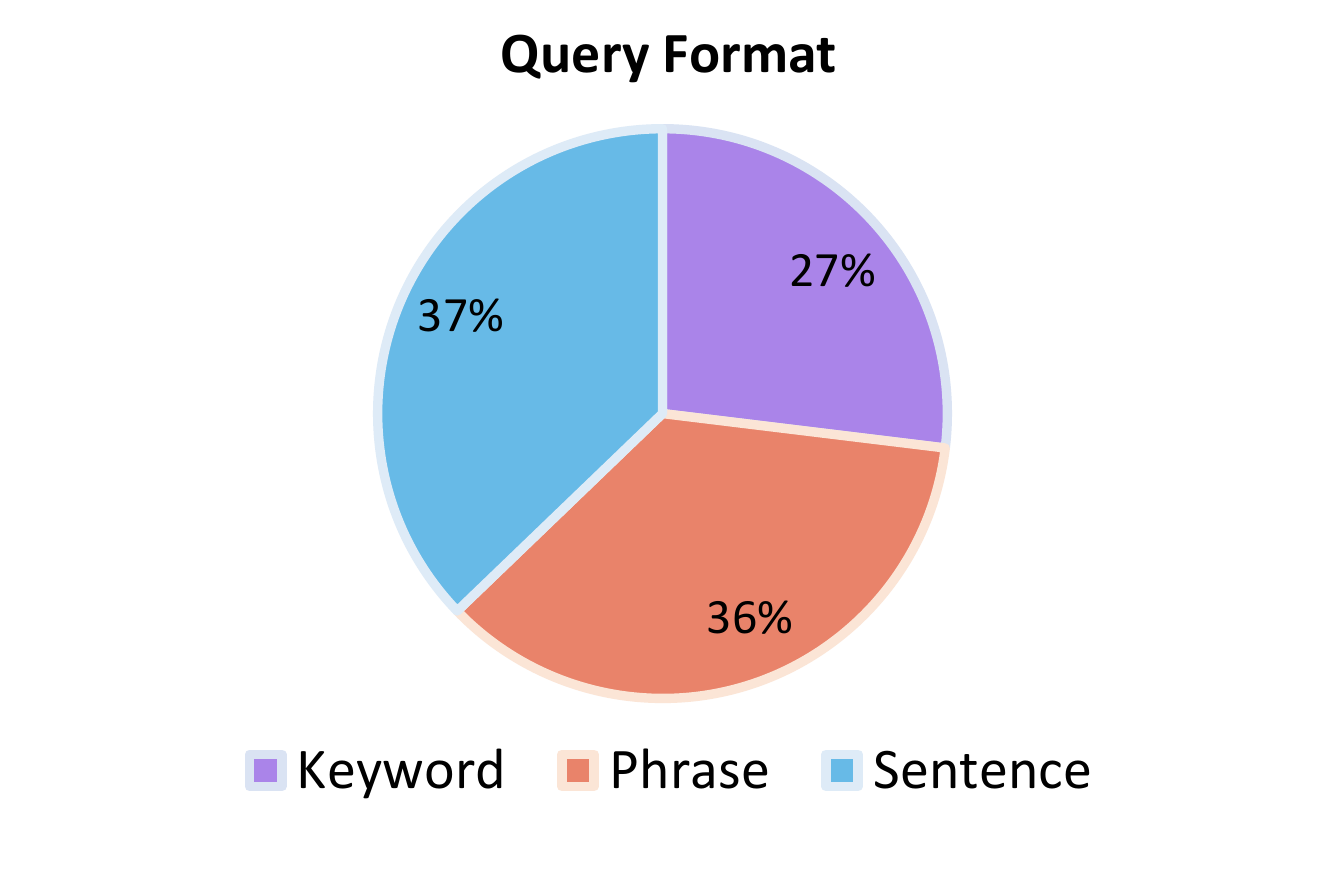}
    \vspace{-0.42cm}
    \caption{The distribution of query modality and format in the VUE-TR-V2 benchmark.}
    \label{fig:distribution}
\end{figure}


\paragraph{Evaluation Metric:} We follow the previous VUE-TR \cite{team2025vidi} benchmark that uses the AUC (Area Under Curve) score of temporal-dimension precision, recall, and IoU (Intersection over Union) in the evaluation, denoted as $\bar{P}$, $\bar{R}$, $\bar{\text{IoU}}$ in Sec. \ref{sec:experiment}. In particular, $\bar{\text{IoU}}$ is used as the primary metric.

\subsection{VUE-PLOT}
\label{sec:vue_plot}

\begin{table}[t]
\centering
\begin{tabular}{l|ccc|cc}
\toprule[1.5pt]
\multirow{2}{*}{{Duration}} & \multicolumn{3}{c|}{\textbf{Character}} & \multicolumn{2}{c}{\textbf{Reasoning}} \\
 & \# Videos & \# Segments & \# Bboxes & \# Videos & \# QAs \\
\hline
$<$90s & 34 & 547 & 1276 & 7 & 65 \\
90-150s & 248 & 5630 & 13422 & 56 & 521 \\
150-210s & 235 & 6517 & 15944 & 58 & 488 \\
$>$210s & 29 & 860 & 2441 & 16 & 140 \\
\hline
Total & 546 & 13554 & 33083 & 137 & 1214 \\
\bottomrule[1.5pt]
\end{tabular}
\caption{
\revise{
Data distribution statistics for the VUE-PLOT benchmark. The benchmark covers a diverse range of video durations, from short clips ($<90$s) to longer sequences ($>210$s), with a total video duration of 22.45 hours for the \textit{Character} track and 6.11 hours for the \textit{Reasoning} track. This comprehensive coverage enables robust evaluation of models across various temporal scales. The \textit{Character} track provides fine-grained grounding annotations (segments and bounding boxes), while the \textit{Reasoning} track challenges models with complex reasoning tasks across multiple domains.}}
\label{tab:VUE_plot_dist}
\end{table}

\begin{figure}[!htbp]
    \centering
    \includegraphics[width=0.9\linewidth]{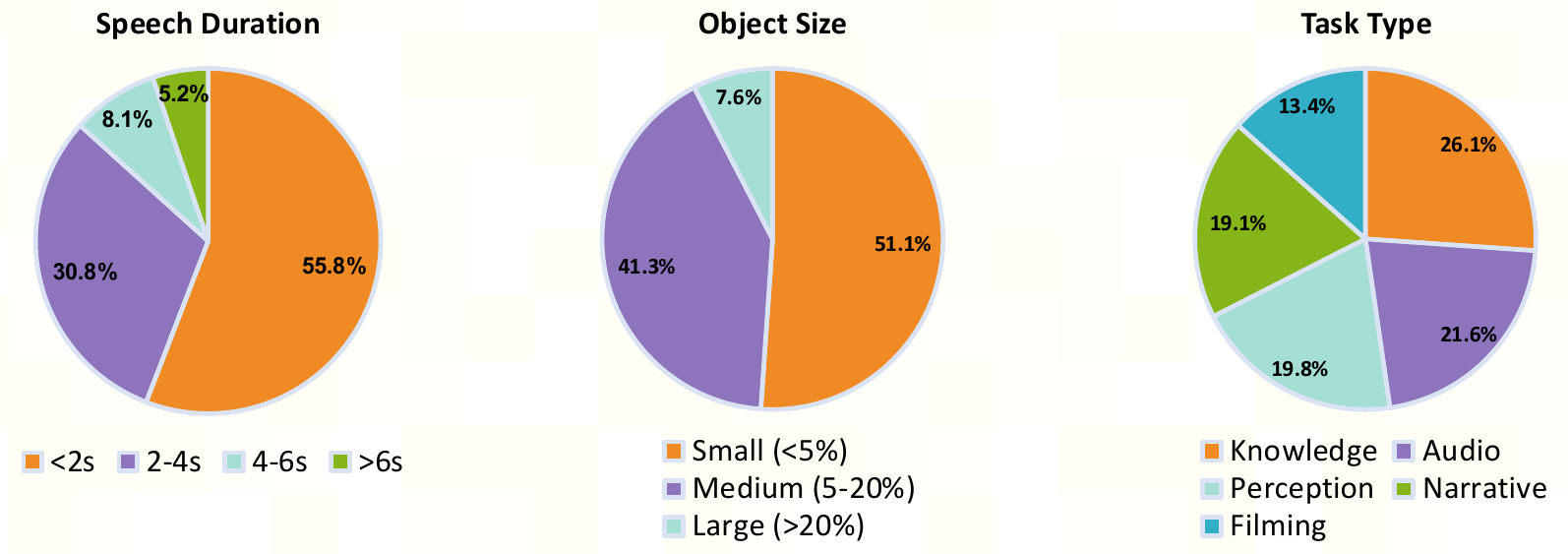}
    \caption{The speech duration distribution (left) and the bounding box area distribution (middle) of the \textit{Character} track of the VUE-PLOT benchmark and task type distribution (right) of the \textit{Reasoning} track.}
    \label{fig:VUE_plot_dist}
\end{figure}
\revise{
The \textit{Character} track of VUE-PLOT serves as the foundation for fine-grained video understanding, comprising $546$ videos with a total duration of $22.45$ hours. 
As detailed in Tab. \ref{tab:VUE_plot_dist}, the videos are categorized into four duration intervals to ensure temporal diversity: $<90$s, $90$-$150$s, $150$-$210$s, and $>210$s. 
The distribution is well-balanced across the middle intervals, providing a solid basis for training and evaluation. 
Beyond video-level stats, the benchmark is densely annotated with $13,554$ speech segments and $33,083$ bounding boxes. 
The segment durations are predominantly short, as shown in Fig. \ref{fig:VUE_plot_dist} (left), requiring precise temporal localization. Meanwhile, the bounding boxes cover a range of object sizes, with $51.1\%$ Small ($<5\%$ area), $41.3\%$ Medium ($5\%$-$20\%$ area), and $7.6\%$ Large ($>20\%$ area) as shown in Fig. \ref{fig:VUE_plot_dist} (middle), challenging models to handle scale variation.
}

\revise{
The \textit{reasoning} track of the VUE-PLOT benchmark, designed for high-level reasoning, consists of $1,214$ QA pairs derived from a diverse subset of $137$ videos (totaling $6.11$ hours). 
The questions are distributed across five distinct task types, as shown in Fig. \ref{fig:VUE_plot_dist} (right): \textit{Narrative and Structural Understanding}, \textit{Perception and Understanding}, \textit{Professional Filming and Editing Techniques}, \textit{Social Cognition and Knowledge Integration}, \textit{Speech, Audio, and Sound Effect Reasoning}. 
This diversity ensures that models are tested on various cognitive dimensions. 
The textual complexity is also significant, with an average question length of $24.8$ words, an average option length of $24.5$ words, necessitating strong multimodal reasoning capabilities.}

\revise
{
\paragraph{Evaluation Metric:} We evaluate the performance of models on plot understanding and reasoning using different metrics.
{\noindent \textbf{Temporal Grounding IoU (\texorpdfstring{$\mathrm{tIoU}$}{tIoU})}} is used to evaluate the temporal alignment accuracy between predicted and ground-truth transcript segments, we match each ground-truth segment including the interested speaking characters to at most one predicted segment based on temporal Intersection-over-Union (IoU). The temporal grounding score is reported as the average IoU over all matched segment pairs:
\[
    \mathrm{tIoU}
    = \frac{1}{N_{\mathrm{match}}} \sum_{i=1}^{N_{\mathrm{match}}}
    \mathrm{IoU}(s_i^{\mathrm{gt}}, s_i^{\mathrm{pred}}).
\]
}

\revise{
{\noindent \textbf{Word Error Rate (\texorpdfstring{$\mathrm{WER}$}{WER})}} measures the speech transcription quality under correct temporal grounding. Specifically, we concatenate the text from all matched ground-truth segments and their corresponding predictions, and compute the Word Error Rate (WER) between the resulting corpora:
\[
    \mathrm{WER}
    = \frac{S + D + I}{N},
\]
where $S$, $D$, and $I$ denote the numbers of substitutions, deletions, and insertions, respectively, and $N$ is the number of words in the ground-truth transcript.
}

\revise{
{\noindent \textbf{Bounding Box IoU.}}
For matched transcript segments that include spatial annotations, we evaluate visual grounding accuracy using bounding box IoU. Ground-truth and predicted boxes are temporally aligned based on minimal timestamp difference (\ie, $20$ms in our setting), and IoU is computed for each matched box pair. The final score is reported as the average bounding box IoU over all matched boxes:
\[
    \mathrm{sIoU}
    = \frac{1}{N_{\mathrm{box}}} \sum_{j=1}^{N_{\mathrm{box}}}
    \mathrm{IoU}(b_j^{\mathrm{gt}}, b_j^{\mathrm{pred}}).
\]
}

\revise{
{\noindent \textbf{Multiple-Choice VQA Accuracy.}}
To evaluate high-level plot understanding in a multiple-choice setting, we measure accuracy on multiple-choice visual question answering tasks. For each question, the model selects one answer from a fixed set of candidate options. The Multiple-Choice VQA Accuracy is defined as the proportion of questions for which the predicted answer exactly matches the ground-truth answer:
\[
    \mathrm{Acc}_{\mathrm{MC}}
    = \frac{1}{N_{\mathrm{q}}} \sum_{i=1}^{N_{\mathrm{q}}}
    \mathbb{I}\!\left( a_i^{\mathrm{pred}} = a_i^{\mathrm{gt}} \right),
\]
where $N_{\mathrm{q}}$ denotes the total number of multiple-choice questions, $a_i^{\mathrm{pred}}$ and $a_i^{\mathrm{gt}}$ represent the predicted and ground-truth answers for the $i$-th question, and $\mathbb{I}(\cdot)$ is the indicator function.
}

\section{Experiment}
\label{sec:experiment}
To validate the effectiveness of \method{}, we evaluate it on three representative tasks: Temporal Retrieval, Spatio-Temporal Grounding, and Video Question Answering.
While the first two are tested on our proposed benchmarks (VUE-TR-V2 and VUE-STG), the latter leverages public video QA datasets \cite{wang2025lvbench,wu2024longvideobench,fu2025video} to assess general multimodal understanding. \revise{VUE-PLOT is added to evaluate \method{} and competitive models for complex plot understanding.}

\subsection{Spatio-Temporal Grounding}

We compare our \method{} model against leading large multimodal models (LMMs), including Gemini 3, GPT-5 and Qwen3-VL, on the VUE-STG benchmark. Since our system predicts full spatio-temporal tubes, we conduct a comprehensive analysis across both temporal and spatial dimensions. 
The metrics follow the definitions in Section~\ref{sssec:stvg_metrics}.

Overall, \method{} achieves the strongest performance across all temporal and spatio-temporal metrics, consistently outperforming Gemini 3, GPT-5 and Qwen3-VL. 
This demonstrates that \method{} not only produces more accurate temporal grounding but also achieves substantially better spatial localization throughout the video sequence.

\subsubsection{Inference Setup for LMMs}
\label{sssec:llm_inference_setup}
To ensure a fair comparison with other LMMs, we carefully design task-aligned prompts and standardized output formats so that all models can perform spatio-temporal grounding in a one-shot setting.
Our \method{} model outputs spatio-temporal tubes as a sequence of discrete, normalized bounding boxes per timestamp,
while other LMMs generate temporal predictions in varying forms according to their APIs.
In the following, we detail the input, prompt, and output settings for each model, and describe how all predictions are converted into a unified temporal representation for consistent evaluation. 

\paragraph{Gemini 3 Input/Output Format:}
We evaluate the Gemini 3 Pro Preview model, which accepts video input via URL. Specifically, we design a structured prompt that instructs the model to output a JSON array containing timestamps and bounding boxes, \ie, 

\begin{quote}
\small
\begin{verbatim}
Answer with timestamps and bounding boxes and do not output explanation.
Output only a JSON array.
Timestamp format: MM:SS with zero-padding (00-59 for MM and SS).
Bounding box format: [x0, y0, x1, y1] with values in [0, 1000] normalized to the video frame.

Example:
[{"timestamp":"00:30", "box_2d":[100, 200, 300, 400]},
 {"timestamp":"05:00", "box_2d":[150, 250, 350, 450]}]

What are all the timestamps and positions corresponding to the text query:
"yellow school bus which parked on street near the tree with yellow leaves"?
\end{verbatim}
\end{quote}

In constructing our prompts, we follow the official Gemini API guidelines for temporal 
and spatial annotations: the \texttt{MM:SS} timestamp format is based on the Video 
Understanding specification~\footnote{\url{https://ai.google.dev/gemini-api/docs/video-understanding}}, 
while the bounding box coordinate system in $[0,1000]$ is defined in the Image 
Understanding documentation~\footnote{\url{https://ai.google.dev/gemini-api/docs/image-understanding}}. 

\paragraph{GPT-5 Input/Output Format:}
GPT-5 API does not accept video directly; instead, it supports a sequence of up to $120$ image frames per request. We sample frames according to the following rule:  
1) for videos shorter than two minutes, we extract frames at $1$ FPS, matching \method{}'s input configuration;  
2) for longer videos, we uniformly subsample frames to satisfy the $120$-frame limit. The prompt for GPT-5 is
\begin{quote}
\small
\begin{verbatim}
Given the frames of video, please find all the objects corresponding to the text query:
"yellow school bus which parked on street near the tree with yellow leaves".

Output only a JSON array.
Frame index format: 0-based integer.
Bounding box format: [x0, y0, x1, y1] with normalized value in 0.000 ~ 1.000

Example:
[{"frame": 3, "box": [0.051, 0.252, 0.323, 0.954]}, 
{"frame": 5, "box": [0.372, 0.353, 0.634, 0.955]}].
\end{verbatim}
\end{quote}

The model outputs bounding boxes defined on sampled frame indices.

\begin{table}[!htbp]
\centering
\resizebox{\linewidth}{!}{
\begin{tabular}{l| l c | c c c c c}
\toprule[1.2pt]
Major Type & Category & Metric(\%) & \textbf{Vidi2.5} & Vidi2 & \makecell{Gemini 3\\Pro Prev.} & GPT-5  & \makecell{Qwen3-VL\\-32B} \\
\hline

\multirow{3}{*}{Overall} & \multirow{3}{*}{Overall} 
~ &   tIoU & \textbf{58.34} & 53.19 & 27.50 & 16.40 & 25.91 \\
~ & ~ & tP & 72.14 & \textbf{73.00} & 51.91 & 38.29 & 45.29 \\
~ & ~ & tR & \textbf{68.00} & 59.80 & 35.26 & 19.53 & 39.19 \\

\hline\hline

\multirow{9}{*}{Video Length} & \multirow{3}{*}{\makecell[l]{Ultra-Short \\ ($<1m$)} } 
~ & tIoU & \textbf{73.13} & 61.43 & 38.68 & 64.66 & 54.33 \\
~ & ~ & tP & \textbf{85.99} & 85.04 & 58.91 & 78.34 & 76.54 \\
~ & ~ & tR & \textbf{79.89} & 66.80 & 52.29 & 75.80 & 69.22 \\
\cline{2-8}
~ & \multirow{3}{*}{\makecell[l]{Short \\ ($1-10m$)}} 
~ & tIoU & \textbf{65.31} & 61.48 & 35.39 & 19.73 & 38.97 \\
~ & ~ & tP & 79.40 & \textbf{81.83} & 66.49 & 43.86 & 58.66 \\
~ & ~ & tR & \textbf{74.04} & 68.36 & 43.59 & 24.27 & 57.27 \\
\cline{2-8}
~ & \multirow{3}{*}{\makecell[l]{Medium \\ ($10-30m$)} } 
~ & tIoU & \textbf{51.63} & 47.27 & 21.13 & 4.10 & 13.19 \\
~ & ~ & tP & 65.48 & \textbf{65.98} & 43.15 & 23.86 & 30.45 \\
~ & ~ & tR & \textbf{62.38} & 54.01 & 27.37 & 4.77 & 23.64 \\
\hline\hline
\multirow{9}{*}{Tube Duration} & \multirow{3}{*}{\makecell[l]{Micro-Short \\ ($<3s$)}} 
~ & tIoU & \textbf{53.40} & 47.24 & 25.92 & 16.70 & 24.44 \\
~ & ~ & tP & \textbf{59.48} & 56.49 & 37.06 & 30.61 & 44.74 \\
~ & ~ & tR & \textbf{68.44} & 59.78 & 39.66 & 20.95 & 35.75 \\
\cline{2-8}
~ & \multirow{3}{*}{\makecell[l]{Ultra-Short \\ ($3s\!-\!10s$)}} 
~ & tIoU & \textbf{58.60} & 54.66 & 29.51 & 17.63 & 25.80 \\
~ & ~ & tP & 71.01 & \textbf{73.13} & 51.77 & 38.17 & 46.23 \\
~ & ~ & tR & \textbf{68.25} & 60.85 & 38.03 & 21.22 & 40.21 \\
\cline{2-8}
~ & \multirow{3}{*}{\makecell[l]{Short \\ ($10s\!-\!60s$)}} 
~ & tIoU & \textbf{59.84} & 52.14 & 23.20 & 13.23 & 26.85 \\
~ & ~ & tP & \textbf{80.49} & 79.93 & 58.71 & 41.25 & 43.23 \\
~ & ~ & tR & \textbf{67.17} & 57.18 & 26.43 & 14.72 & 38.18 \\
\hline\hline
\multirow{9}{*}{Object Size} & \multirow{3}{*}{\makecell[l]{Small \\ ($<10\%$)} }
~ & tIoU & \textbf{56.35} & 51.46 & 28.35 & 17.57 & 26.01 \\
~ & ~ & tP & \textbf{71.05} & 70.79 & 52.00 & 40.94 & 46.46 \\
~ & ~ & tR & \textbf{65.55} & 57.94 & 37.16 & 20.46 & 38.99 \\

\cline{2-8}
~ & \multirow{3}{*}{\makecell[l]{Medium \\ ($10\%-30\%$)}}
~ & tIoU & \textbf{60.98} & 54.51 & 27.23 & 16.66 & 26.85 \\
~ & ~ & tP & \textbf{73.98} & 73.85 & 51.58 & 36.91 & 45.04 \\
~ & ~ & tR & \textbf{71.15} & 61.56 & 34.41 & 20.04 & 40.76 \\

\cline{2-8}
~ & \multirow{3}{*}{\makecell[l]{Large \\ ($>30\%$)}}
~ & tIoU & \textbf{58.72} & 54.97 & 26.06 & 13.49 & 24.32 \\
~ & ~ & tP & 71.78 & \textbf{76.55} & 52.23 & 34.77 & 43.18 \\
~ & ~ & tR & \textbf{68.62} & 61.21 & 32.38 & 16.76 & 37.29 \\
\bottomrule[1.2pt]
\end{tabular}}

\caption{
Temporal grounding results on the VUE-STG benchmark, reporting temporal precision (tP), recall (tR), and intersection-over-union (tIoU). tR and tIoU are averaged over ground-truth tubes, while tP is averaged over predicted tubes. Results are further grouped by video length, annotated tube duration, and object size.}
\label{tab:stvg_results_t}
\end{table}

\begin{table}[!htbp]
\centering
\resizebox{\linewidth}{!}{
\begin{tabular}{l| l c | c c c c c}
\toprule[1.2pt]
Major Type & Category & Metric(\%) & \textbf{Vidi2.5} & Vidi2 & \makecell{Gemini 3\\Pro Prev.} & GPT-5  & \makecell{Qwen3-VL\\-32B} \\
\hline

\multirow{4}{*}{Overall} & \multirow{4}{*}{Overall}
~ & vIoU & \textbf{38.64} & 32.57 & 4.61 & 5.47 & 5.12 \\
~ & ~ & vIoU-Int. & \textbf{64.84} & 60.30 & 16.59 & 33.64 & 18.47 \\
~ & ~ & vP & \textbf{47.26} & 44.56 & 8.95 & 13.01 & 8.61 \\
~ & ~ & vR & \textbf{44.71} & 36.32 & 5.71 & 6.50 & 7.49 \\
\hline\hline
\multirow{12}{*}{Video Length} & \multirow{4}{*}{\makecell[l]{Ultra-Short \\ ($<1m$)} }
~ & vIoU & \textbf{48.25} & 36.70 & 7.77 & 20.60 & 11.30 \\
~ & ~ & vIoU-Int. & \textbf{64.42} & 57.86 & 19.44 & 31.36 & 20.56 \\
~ & ~ & vP & \textbf{56.17} & 50.06 & 12.44 & 25.19 & 16.11 \\
~ & ~ & vR & \textbf{52.05} & 39.46 & 9.96 & 23.98 & 13.95 \\

\cline{2-8}
~ & \multirow{4}{*}{\makecell[l]{Short \\ ($1-10m$)}}
~ & vIoU & \textbf{45.23} & 39.62 & 6.36 & 6.86 & 8.63 \\
~ & ~ & vIoU-Int. & \textbf{67.78} & 62.97 & 18.45 & 34.45 & 21.01 \\
~ & ~ & vP & \textbf{54.68} & 52.41 & 12.36 & 14.97 & 12.37 \\
~ & ~ & vR & \textbf{50.60} & 43.45 & 7.61 & 8.43 & 12.56 \\

\cline{2-8}
~ & \multirow{4}{*}{\makecell[l]{Medium \\ ($10-30m$)} }
~ & vIoU & \textbf{33.27} & 28.18 & 3.05 & 1.44 & 2.02 \\
~ & ~ & vIoU-Int. & \textbf{63.29} & 59.37 & 14.33 & 34.52 & 15.08 \\
~ & ~ & vP & \textbf{41.63} & 39.46 & 6.49 & 8.47 & 4.75 \\
~ & ~ & vR & \textbf{40.18} & 32.10 & 3.84 & 1.68 & 3.55 \\
\hline\hline
\multirow{12}{*}{Tube Duration} & \multirow{4}{*}{\makecell[l]{Micro-Short \\ ($<3s$)}}
~ & vIoU & \textbf{33.35} & 27.96 & 3.93 & 5.60 & 4.31 \\
~ & ~ & vIoU-Int. & \textbf{59.05} & 57.28 & 12.98 & 32.95 & 16.06 \\
~ & ~ & vP & \textbf{36.98} & 33.25 & 5.85 & 10.52 & 8.10 \\
~ & ~ & vR & \textbf{41.31} & 34.74 & 5.20 & 6.74 & 5.34 \\
\cline{2-8}
~ & \multirow{4}{*}{\makecell[l]{Ultra-Short \\ ($3s\!-\!10s$)}}
~ & vIoU & \textbf{39.22} & 34.01 & 4.79 & 5.74 & 4.57 \\
~ & ~ & vIoU-Int. & \textbf{65.73} & 60.98 & 16.30 & 33.40 & 17.32 \\
~ & ~ & vP & \textbf{47.10} & 45.17 & 8.66 & 12.90 & 8.07 \\
~ & ~ & vR & \textbf{45.32} & 37.52 & 6.08 & 6.94 & 7.15 \\

\cline{2-8}
~ & \multirow{4}{*}{\makecell[l]{Short \\ ($10s\!-\!60s$)}}
~ & vIoU & \textbf{39.54} & 31.03 & 4.48 & 4.76 & 6.85 \\
~ & ~ & vIoU-Int. & \textbf{64.99} & 59.83 & 18.43 & 34.27 & 22.45 \\
~ & ~ & vP & \textbf{52.19} & 48.00 & 11.03 & 14.14 & 10.16 \\
~ & ~ & vR & \textbf{44.67} & 34.01 & 5.01 & 5.29 & 9.27 \\

\hline\hline
\multirow{12}{*}{Object Size} & \multirow{4}{*}{\makecell[l]{Small \\ ($<10\%$)} }
~ & vIoU & \textbf{29.08} & 23.31 & 2.33 & 3.66 & 1.49 \\
~ & ~ & vIoU-Int. & \textbf{49.86} & 44.13 & 7.90 & 21.38 & 5.88 \\
~ & ~ & vP & \textbf{36.20} & 32.08 & 4.36 & 9.06 & 2.73 \\
~ & ~ & vR & \textbf{33.25} & 25.71 & 2.85 & 4.18 & 2.17 \\

\cline{2-8}
~ & \multirow{4}{*}{\makecell[l]{Medium \\ ($10\%-30\%$)}}
~ & vIoU & \textbf{46.08} & 39.42 & 5.83 & 7.34 & 7.08 \\
~ & ~ & vIoU-Int. & \textbf{73.99} & 71.21 & 20.41 & 42.19 & 23.65 \\
~ & ~ & vP & \textbf{55.33} & 52.97 & 10.92 & 15.93 & 11.29 \\
~ & ~ & vR & \textbf{53.37} & 44.44 & 7.16 & 8.69 & 10.20 \\

\cline{2-8}
~ & \multirow{4}{*}{\makecell[l]{Large \\ ($>30\%$)}}
~ & vIoU & \textbf{48.38} & 42.50 & 7.77 & 6.61 & 10.09 \\
~ & ~ & vIoU-Int. & \textbf{82.82} & 77.22 & 30.22 & 48.08 & 38.02 \\
~ & ~ & vP & \textbf{59.32} & 59.14 & 15.86 & 16.96 & 17.08 \\
~ & ~ & vR & \textbf{56.74} & 47.29 & 9.78 & 8.27 & 15.00 \\
\bottomrule[1.2pt]
\end{tabular}}

\caption{
Spatio-temporal grounding results on the VUE-STG benchmark, reporting spatio-temporal precision (vP), recall (vR), and intersection-over-union (vIoU and vIoU-Int.).
vR and vIoU are averaged over ground-truth tubes, while vP is averaged over predicted tubes. vIoU-Int. is computed only over the predicted tubes that overlap with a ground-truth tube. Results are grouped by video length, annotated tube duration, and object size.
}
\label{tab:stvg_results_v}
\end{table}

\paragraph{Qwen3-VL Input/Output Format:}
For Qwen3-VL, we test the \texttt{Qwen3-VL-32B-Instruct} model following the spatio-temporal grounding examples provided in the Qwen3-VL
blog\footnote{\url{https://qwen.ai/blog?id=qwen3-vl}} and the interleaved
timestamp-image pattern demonstrated in the official cookbook\footnote{\url{https://github.com/QwenLM/Qwen3-VL/blob/main/cookbooks/video_understanding.ipynb}}.

Following the conventions illustrated by the cookbook, each video is converted into a sequence of sampled frames, and the prompt is formed by interleaving explicit timestamp indicators (\eg, ``\texttt{<0.0 seconds>}''), the corresponding frame image, the next timestamp, and so on. 
After all timestamp-image pairs are provided, the textual grounding query is appended, \eg,
\begin{quote}
\small
\begin{verbatim}
Given the query "yellow school bus which parked on street near the tree with 
yellow leaves", for each frame, detect and localize the visual content described 
by the given textual query in JSON format. If the visual content does not exist in
a frame, skip that frame.

Output Format: [{"time": 1.0, "bbox_2d": [x_min, y_min, x_max, y_max]}, 
{"time": 2.0, "bbox_2d": [x_min, y_min, x_max, y_max]}, ...].
\end{verbatim}
\end{quote}
Qwen3-VL outputs timestamps directly in seconds, and its bounding box coordinates are defined over an integer range of $[0,1000]$, which coincides with the range used by the Gemini API.

\paragraph{Temporal and Spatial Normalization:}
The four models generate temporal grounding signals in mutually incompatible representations. To enable fair comparison, we convert all predictions into a unified timeline measured in seconds. Specifically, 
1) Gemini 3's \texttt{MM:SS} timestamps are parsed and converted into seconds;  
2) GPT-5's frame indices are mapped to seconds according to the sampling policy;  
3) Qwen3-VL directly reports timestamps in seconds, which does not require additional temporal transformation.  
After conversion, all timestamps are discretized by rounding to the nearest
integer second to obtain a consistent per-second temporal representation.

For spatial normalization, we map all bounding boxes to a shared coordinate
system in $[0,1]$. Gemini and Qwen3-VL use an integer-based coordinate range of $[0,1000]$, which we normalize by dividing by $1000$. GPT-5 already produces normalized coordinates.
This yields a unified spatial representation from which all spatial and spatio-temporal metrics are computed.

\subsubsection{Performance Evaluation}
\paragraph{Temporal Performance:}
Table~\ref{tab:stvg_results_t} summarizes the temporal grounding results across different video lengths, 
ground-truth temporal durations, and object sizes. 
\method{} establishes a clear advantage in all conditions, achieving the highest  temporal precision ($73.00$), recall ($59.80$), and IoU ($53.19$).  
The gap is especially pronounced on medium-length videos, 
where Gemini 3 Pro Preview, GPT-5 and Qwen3-VL-32B degrade significantly. 
These findings highlight that \method{} maintains robust temporal alignment even under long-duration or fine-grained temporal conditions.

\paragraph{Spatio-Temporal Performance:}
Table~\ref{tab:stvg_results_v} reports the spatio-temporal grounding accuracy. 
Across all categories, \method{} again delivers the strongest results, with the highest vP ($44.56$), vR ($36.32$), and vIoU ($32.57$), as well as a large margin on the intersection-based metric vIoU-Int ($60.30$). 
These improvements indicate that \method{} not only identifies the correct temporal extent but also maintains coherent spatial localization throughout the entire tube. \methodnew{} further achieves significant improvement (6.07\%) over \method{} on vIoU with additional RL training. 
In contrast, Gemini 3 Pro Preview, GPT-5 and Qwen3-VL-32B exhibit severe degradation on long videos or small-object scenarios, suggesting instability in spatial tracking when reasoning over extended temporal horizons. 

\begin{table}[!htbp]
    \centering
    \begin{tabular}{l c | c c c c }
    \toprule[1.2pt]
       Category & Metric(\%) & \textbf{Vidi2.5} & \method{}& \makecell{Gemini 3\\ Pro Prev.}  & GPT-5 \\
        \hline
        \multirow{3}{*}{Overall}
        ~ & $\bar{\text{IoU}}$ & \textbf{49.62} & 48.75 & 37.58 & 17.15 \\
        ~ & $\bar{\text{P}}$ & 59.78 & \textbf{62.45} & 48.61 & 29.64 \\
        ~ & $\bar{\text{R}}$ & \textbf{71.09} & 64.93 & 56.30 & 26.63 \\
        \hline\hline

        \multirow{3}{*}{Ultra-Short ($<1m$)}
        ~ & $\bar{\text{IoU}}$ & 57.60 & \textbf{59.09} & 56.72 & 41.08 \\
        ~ & $\bar{\text{P}}$ & 66.81 & \textbf{69.76} & 68.56 & 63.84 \\
        ~ & $\bar{\text{R}}$ & 80.66 & \textbf{82.48}  & 75.22 & 48.19 \\

        \hline
        \multirow{3}{*}{Short ($1-10m$)}
        ~ & $\bar{\text{IoU}}$ & \textbf{50.63} & 49.01 & 49.40 & 26.52 \\
        ~ & $\bar{\text{P}}$ & 63.61 & \textbf{65.38} & 61.83 & 41.64 \\
        ~ & $\bar{\text{R}}$ & 73.31 & 66.05 & \textbf{74.93} & 38.91\\

        \hline
        \multirow{3}{*}{Medium ($10-30m$)}
        ~ &  $\bar{\text{IoU}}$ & \textbf{51.06} & 51.05 & 40.56 & 11.28 \\
        ~ & $\bar{\text{P}}$ & 59.94 & \textbf{63.42} & 50.48 & 20.51 \\
        ~ &  $\bar{\text{R}}$ & \textbf{73.55} & 67.91 & 63.69 & 21.50 \\

        \hline
        \multirow{3}{*}{Long ($30-60m$)}
        ~ & $\bar{\text{IoU}}$ & \textbf{48.54} & 48.15 & 26.20 & 9.39 \\        
        ~ & $\bar{\text{P}}$ & 56.96 & \textbf{60.54} & 36.73 & 20.21 \\
        ~ & $\bar{\text{R}}$ & \textbf{68.90} & 62.56 & 39.36 & 16.05 \\

        \hline
        \multirow{3}{*}{Ultra-Long ($>60m$)}
        ~ &  $\bar{\text{IoU}}$ & \textbf{42.22} & 38.65 & 21.19 & 12.49  \\
        ~ & $\bar{\text{P}}$ & 53.86 & \textbf{54.55} & 32.18 & 21.49 \\
        ~ &  $\bar{\text{R}}$ & \textbf{60.73} & 50.98 & 30.11 & 22.27 \\

        \hline\hline
        \multirow{3}{*}{Keyword}
        ~ &  $\bar{\text{IoU}}$ & \textbf{48.56} & 47.73 & 38.41 & 23.17 \\
        ~ & $\bar{\text{P}}$ & 59.64 & \textbf{63.21} & 52.43 & 41.07 \\
        ~ &  $\bar{\text{R}}$ & \textbf{70.63} & 63.72 & 56.00 & 32.59 \\

        \hline
        \multirow{3}{*}{Phrase}
        ~ &  $\bar{\text{IoU}}$ & \textbf{51.22} & 50.11 & 37.84 & 17.02 \\
        ~ & $\bar{\text{P}}$ & 61.66 & \textbf{63.80} & 48.94 & 29.18 \\
        ~ &  $\bar{\text{R}}$ & \textbf{73.10} & 67.07 &  57.40 & 25.88 \\

        \hline
        \multirow{3}{*}{Sentence} 
        ~ &  $\bar{\text{IoU}}$ & \textbf{48.83} & 48.17 & 36.73 & 12.92 \\
        ~ & $\bar{\text{P}}$ & 58.07 & \textbf{60.59} & 45.51 & 22.26 \\
        ~ &  $\bar{\text{R}}$ & \textbf{69.47} & 63.75 & 55.46 & 23.04 \\
        \hline\hline

        \multirow{3}{*}{Audio}
        ~ &  $\bar{\text{IoU}}$ & 45.64 & \textbf{46.90} & 33.27 & 6.93  \\
        ~ & $\bar{\text{P}}$ & 54.26 & \textbf{57.51} & 39.42 & 12.16 \\
        ~ &  $\bar{\text{R}}$ & \textbf{71.94} & 67.09 & 58.26 & 23.21 \\
        \hline
        \multirow{3}{*}{Vision} 
        ~ &  $\bar{\text{IoU}}$ & \textbf{52.77} & 51.04 & 39.54 & 20.62\\
        ~ & $\bar{\text{P}}$ & 61.37 & \textbf{63.12} & 52.16 & 35.78 \\
        ~ &  $\bar{\text{R}}$ & \textbf{73.13} & 66.94 & 54.76 & 26.85 \\

        \hline
        \multirow{3}{*}{Vision+Audio} 
        ~ &  $\bar{\text{IoU}}$ & \textbf{47.65} & 46.81 & 37.26 & 17.85 \\
        ~ & $\bar{\text{P}}$ & 60.51 & \textbf{64.03} & 48.66 & 30.17 \\
        ~ &  $\bar{\text{R}}$ & \textbf{68.13} & 61.38 & 57.26 & 28.04 \\

    \bottomrule[1.2pt]
    \end{tabular}
    \caption{
    Performance of different models on the VUE-TR-V2 benchmark across various evaluation attributes. $\bar{P}$ and $\bar{R}$ denote the AUC (Area Under Curve) of precision and recall, respectively, while $\bar{\text{IoU}}$ represents the AUC of intersection-over-union between predicted and ground-truth timestamp ranges, following the definition in~\cite{team2025vidi}.
    GPT models are evaluated via the Azure API, and Gemini models via the official Google API. The latest Gemini 3 Pro (Preview) model is tested by directly uploading full videos (no size restriction), whereas GPT-5 is limited by the Azure API's $120$-frame cap; for videos exceeding $120$ seconds, we uniformly sample $120$ frames.}
    \label{tab:tr_results}
\end{table}

\subsection{Temporal Retrieval}
We compare our model with three state-of-the-art proprietary models including Gemini 3 Pro Preview \cite{comanici2025gemini} and GPT-5 \cite{OpenAI2025GPT5SystemCard}. 
There models are chosen for their strong performance and broad modality support, making them competitive baselines for real-world temporal retrieval. \\

\noindent The latest Gemini 3 Pro Preview naturally supports '$\ast$.mp4' as video input with maximum length over $2$ hours. Compared with previous 0325 version, the latest version can follow the temporal retrieval task very well and the result stably contains timestamps in \texttt{HH:MM:SS}. One successful example prompt is used as follows.
\begin{quote}
\texttt{Answer with time ranges and do not output explanation. What are all the time ranges corresponding to the text query: "{QUERY}"?.}
\end{quote}

\noindent Since The latest GPT-5 API does not support audio, we extract frames at $1$ fps and feed them as input images. Following Vidi \cite{team2025vidi}, we design a simple custom prompt with in-context examples, ensuring that the output only contains frame index ranges. An example instruction is shown as
\begin{quote}
\texttt{The input images are frames from a video. Output the frame indexes that correspond to the text query: "{QUERY}". Only output the index range, for example, 2-4, 6-8.}
\end{quote}
We observe that GPT-5 follows this prompt reliably, typically producing clean and parseable frame ranges, which are then converted into time intervals for evaluation. \\

\noindent We present the detailed results in terms of length, query format, and modality subsets in Table \ref{tab:tr_results}. Notably, \method{} significantly outperforms Gemini 3 Pro Preview and GPT-5 for overall IoU ($>10\%$). For detailed analysis on different attributes, \method{} also excels on IoU, precision, recall on most subsets. We can conclude that a higher overall IoU of \method{} score ensures a better experience for automatic video editing/trimming scenarios.\\

\begin{figure}[t]
  \centering
  \captionsetup[subfigure]{labelformat=empty}
  \begin{subfigure}{0.32\textwidth}
    \includegraphics[width=\linewidth]{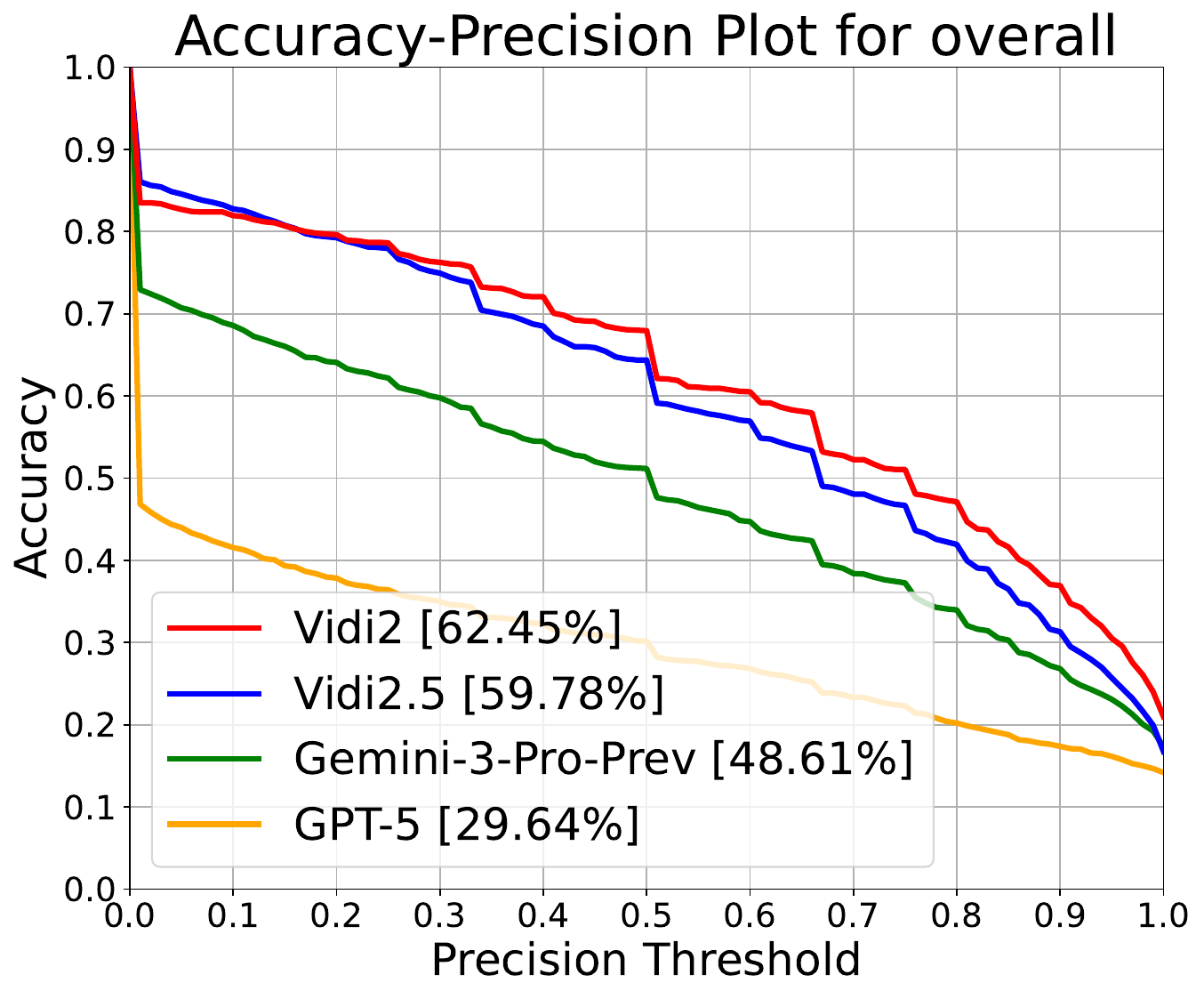}
    \label{fig:sub1}
  \end{subfigure}
  \hfill
  \begin{subfigure}{0.32\textwidth}
    \includegraphics[width=\linewidth]{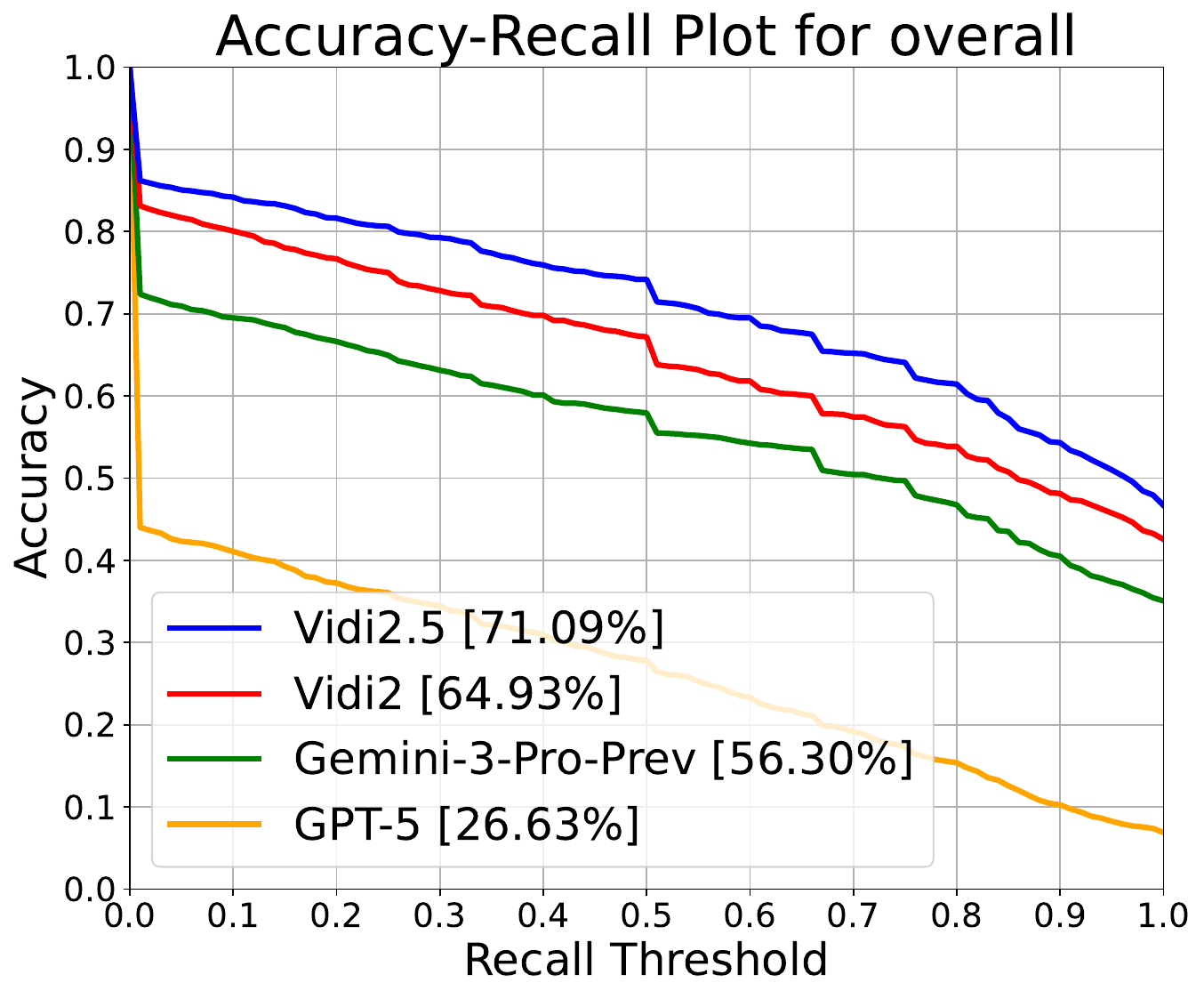}
    \label{fig:sub2}
  \end{subfigure}
  \hfill
  \begin{subfigure}{0.32\textwidth}
    \includegraphics[width=\linewidth]{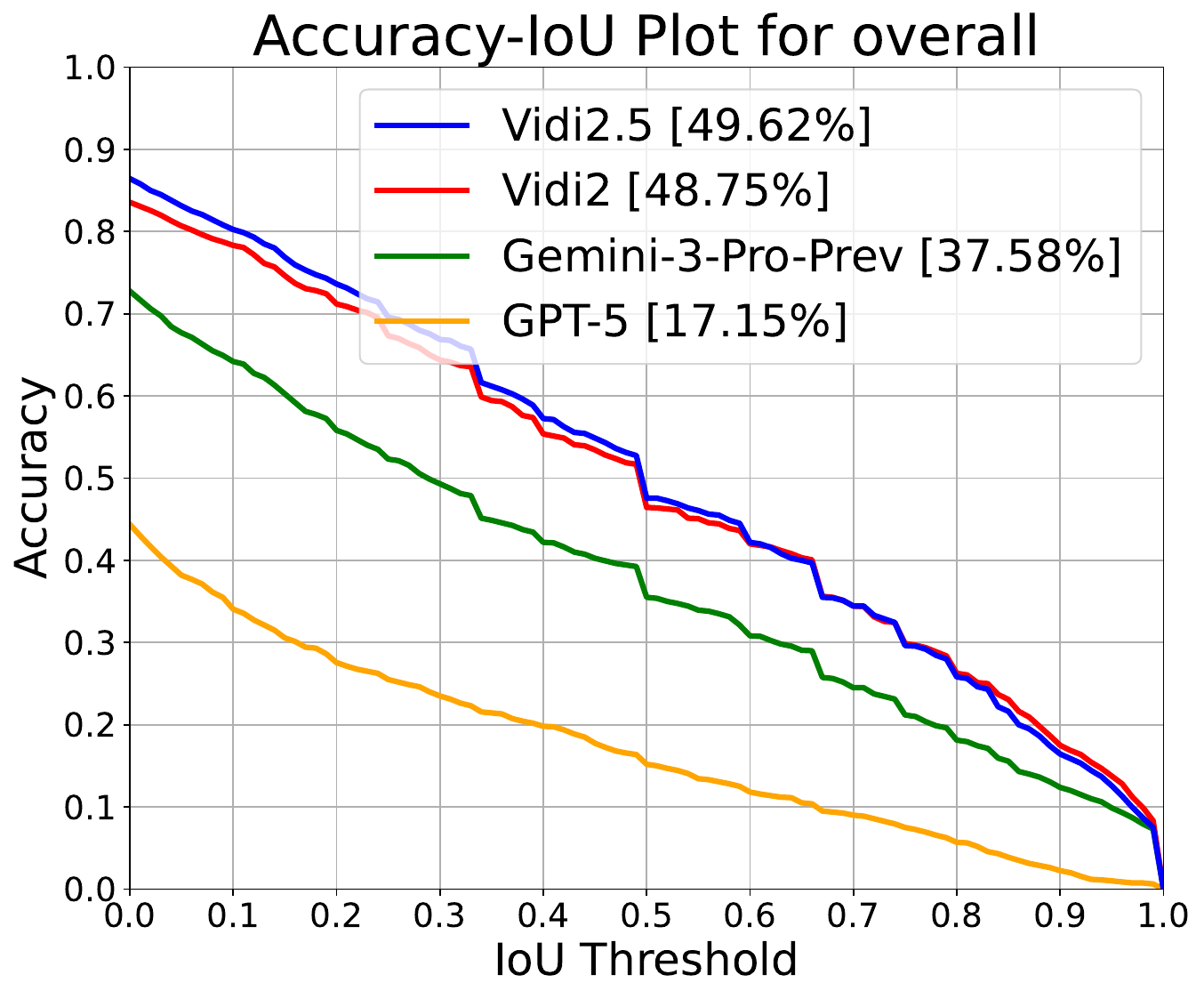}
    \label{fig:sub3}
  \end{subfigure}
  \caption{Overall performance curves for temporal retrieval on the VUE-TR-V2 benchmark. We report accuracy across varying thresholds for different models.}
  \label{fig:curves}
\end{figure}

\noindent We also present the ROC curve for different metrics in Figure \ref{fig:curves}, and we can observe that \method{} consistently outperforms all competitors in IOU, precision and recall across the entire range of thresholds. It shows \method{}'s strength in fine-grained temporal understanding, a critical capability for real-world video editing tasks. \methodnew{} is slightly improved over \method{} on the overall IoU with RL training.

\subsection{Generic Video QA}
As a video understanding foundation model, \method{} also supports video question answering, which is a major improvement over the previous Vidi model.
We evaluate \method{} and several popular multimodal models on three representative video QA benchmarks: LVBench~\cite{wang2025lvbench}, LongVideoBench~\cite{wu2024longvideobench}, and VideoMME~\cite{fu2025video}.
While \method{} still lags behind the leading proprietary model, Gemini-2.5-Pro~\cite{comanici2025gemini}, in general video QA performance, it achieves competitive results compared to the most popular open-source multimodal model of similar scale, namely Qwen2.5-VL-7B~\cite{qwen2.5-VL}.
These results demonstrate that \method{} possesses solid multimodal reasoning ability despite being primarily optimized for retrieval and grounding tasks.

\begin{table}[!htbp]
    \centering
    \begin{tabular}{c|c c c}
    \toprule
         & LVBench \cite{wang2025lvbench} & LongVideoBench \cite{wu2024longvideobench} & VideoMME \cite{fu2025video}\\
    \hline
    Gemini-2.5-Pro \cite{comanici2025gemini} & 78.7 & 84.3 & -\\
     Qwen2.5-VL-7B \cite{qwen2.5-VL}  &  45.3 & 54.7  & 65.1 \\
     \method{} & 45.8 & 57.1 & 63.5 \\
     \methodnew{} & 45.2 & 58.9 & 63.6 \\
     \bottomrule
    \end{tabular}
    \caption{Video QA performance on popular benchmarks.}
    \label{tab:vqa}
\end{table}

\begin{figure*}[!htbp]
    \centering
    \includegraphics[width=\linewidth]{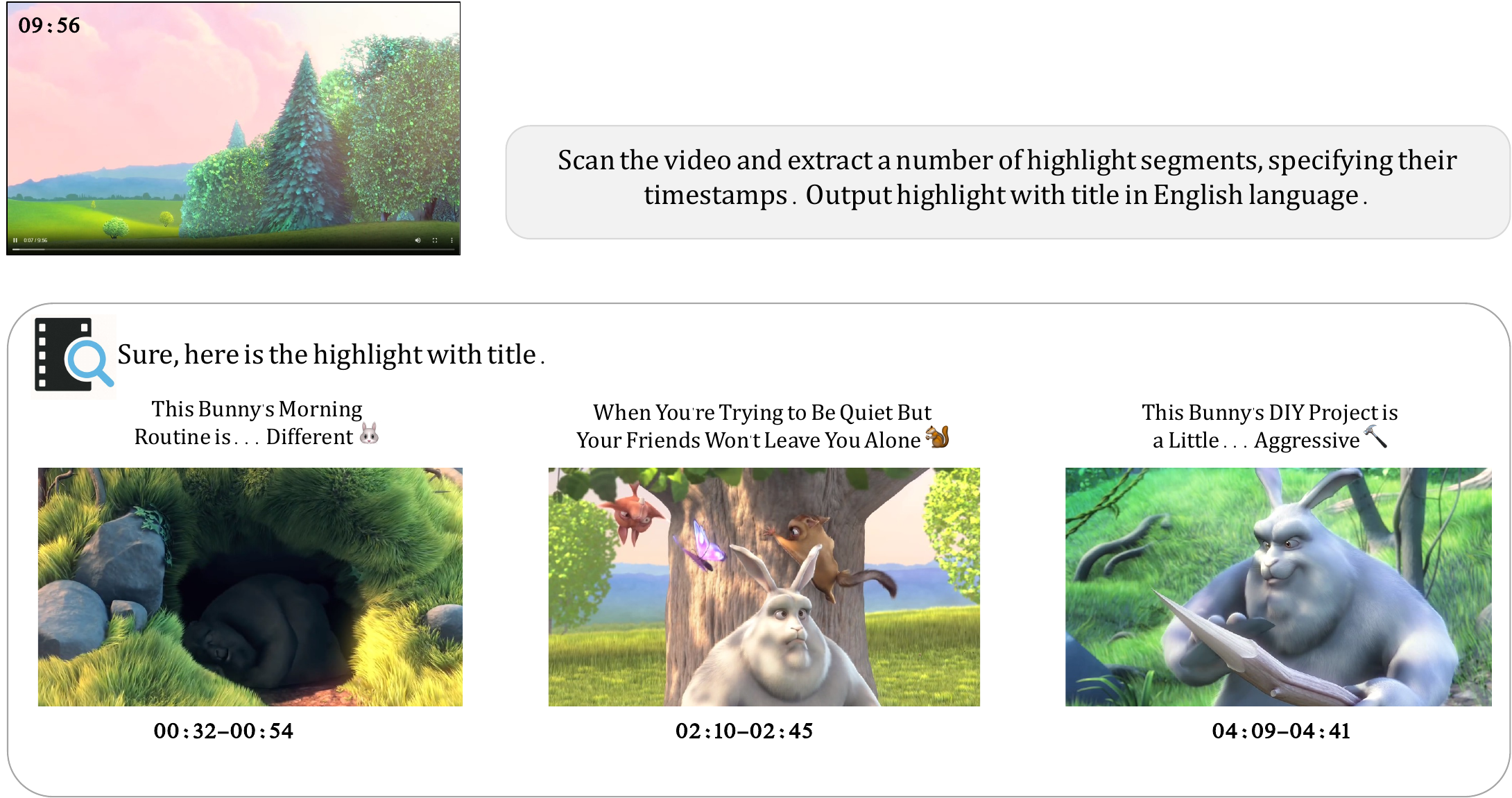}
    \caption{Example of highlight extraction application.}
    \label{fig:application_highlight}
\end{figure*}
\begin{figure*}[!htbp]
    \centering
    \includegraphics[width=\linewidth]{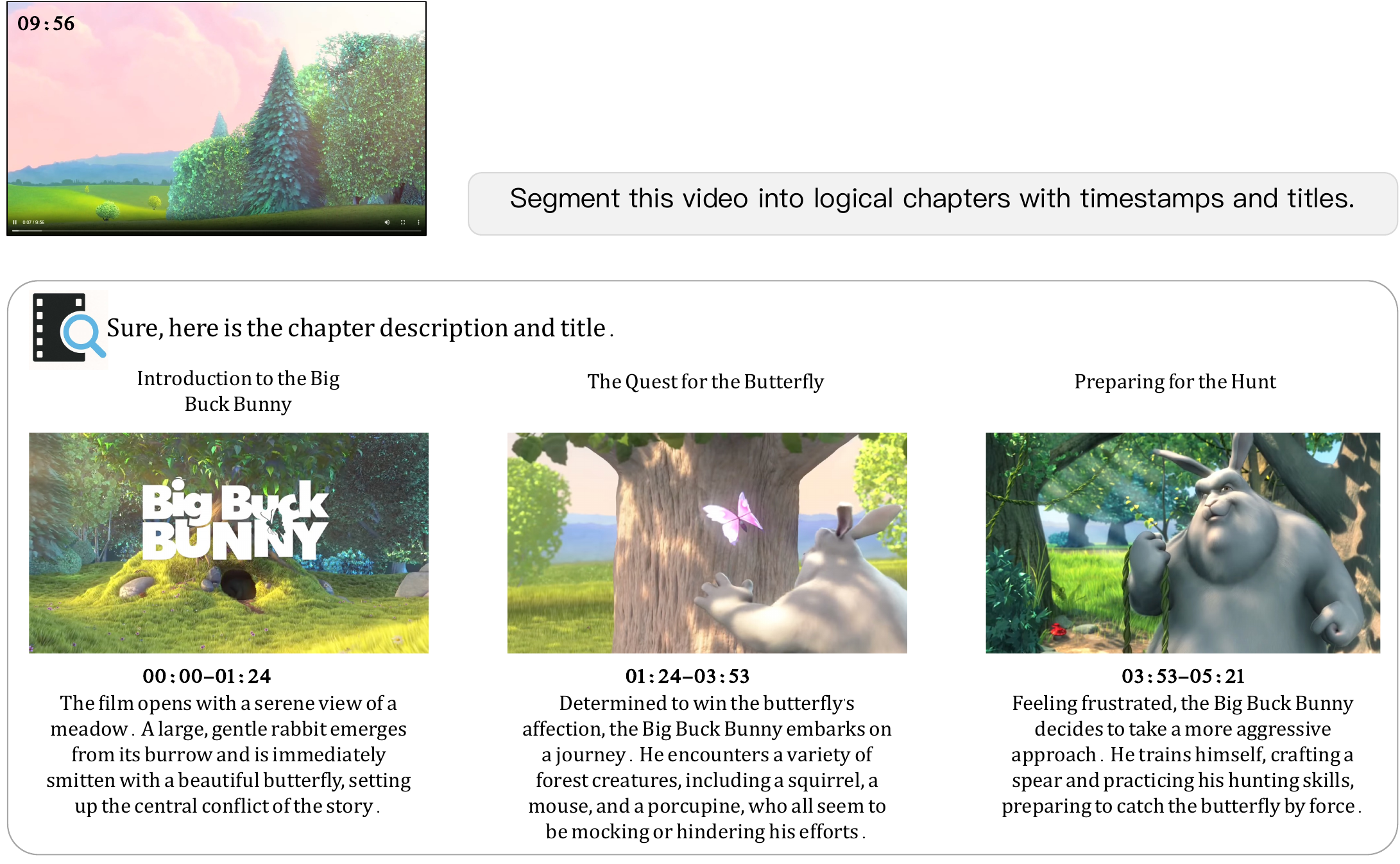}
    \caption{Example of chapter summary application.}
    \label{fig:application_chapter}
\end{figure*}

The strong temporal modeling capability of \method{} enables straightforward adaptation to downstream tasks such as highlight prediction. Given a long raw video, the model can automatically generate multiple highlight segments along with concise titles, allowing users to directly publish the resulting short clips.
As shown in Figure~\ref{fig:application_highlight}, the generated titles align closely with the corresponding video content, demonstrating both temporal precision and semantic understanding. Similarly, as shown in Figure~\ref{fig:application_chapter}, the model can generate chapters with summary and titles for a long raw video.

\subsection{Plot Understanding and Reasoning}
\label{sec:exp_plot}
\textbf{Inference Setup for LMMs.}
For the \textbf{Character} Track of VUE-PLOT, we design task-aligned prompts and structured output formats to encourage the evaluated MLLMs to generate stable and parsable outputs when performing active speaker transcription and face tracking. This design facilitates a fair comparison with \methodnew{}. In the following, we detail the input and output configurations for the compared models.\\

\noindent\textbf{Gemini 3 and Qwen3-Omni Input/Output Format:} We evaluate Gemini 3 Pro Preview \cite{comanici2025gemini} and Qwen3-Omni-Thinking \cite{xu2025qwen3omnitechnicalreport}, leveraging their multimodal capabilities to process both video and audio inputs. For all experiments, the video inputs are sampled at $1$ FPS. To minimize the influence of prompt variations on the evaluation results, we apply the same structured prompt to both models and enforce a unified output format via a JSON schema at the API level. Notably, the structured output functionality of Qwen3-Omni \cite{xu2025qwen3omnitechnicalreport} is supported in the latest version of vLLM\footnote{\url{https://github.com/vllm-project/vllm}}. Under these constraints, both models are required to produce a JSON object containing precise timestamps and normalized coordinates, \ie,
\begin{quote}
\small
\begin{verbatim}
Transcribe the active speaker's speech and track their face
using bounding boxes. Output only a single JSON object and 
do not output explanation.
Timestamp format: Total seconds (float).
Bounding box format: [x0, y0, x1, y1] with values in [0, 1] normalized
to the video frame, tightly enclosing the speaker's face.
Sampling strategy: Sample at the first visible frame and every subsequent
integer second.
Example:
{
  "text": "Hello everyone.",
  "start": 62.4,
  "end": 65.0,
  "boxes": [
    {"timestamp": 62.4, "box_2d": [0.400, 0.150, 0.600, 0.350]},
    {"timestamp": 63.0, "box_2d": [0.405, 0.155, 0.605, 0.355]}
  ]
}
\end{verbatim}
\end{quote}

\begin{table}[thbp]
\centering
\setlength{\tabcolsep}{3pt}
\resizebox{\linewidth}{!}{
\begin{tabular}{l| l| c | c c c c c}
\toprule[1.2pt]
\multirow{3}{*}{Track} & \multirow{3}{*}{Category} & \multirow{3}{*}{\shortstack{Metric\\(\%)}}
& \multicolumn{5}{c}{Models} \\
\cline{4-8}
~ & ~ & ~ 
& \shortstack{\textbf{Vidi}\\\textbf{2.5}}
& \shortstack{Gemini\\3 Pro}
& \shortstack{GPT\\5}
& \shortstack{Qwen3\\VL}
& \shortstack{Qwen3\\Omni} \\
\hline

\multirow{3}{*}{Char.} & \multirow{3}{*}{$\qquad$$\qquad$$\qquad$$\qquad$$\qquad$-}
~ & tIoU & \textbf{71.63} & 66.04 & - & - & 50.68 \\
~ & ~ & sIoU & \textbf{55.89} & 13.24 & - & - & 6.80 \\
~ & ~ & WER $\downarrow$ & \textbf{23.20} & 29.00 & - & - & 58.12 \\

\hline\hline
\multirow{5}{*}{Rea.} 
~ & Perception and Understanding & Acc  & 66.25 & \textbf{80.42} & 55.83 & 35.00 & 40.83 \\
~ & Speech, Audio, and Sound Effect Reasoning & Acc & \textbf{74.43} & 66.03 & 55.34 & 45.42 & 27.10 \\
~ & Social Cognition and Knowledge Integration & Acc & \textbf{62.78} & 62.15 & 53.94 & 31.23 & 23.03 \\
~ & Narrative and Structural Understanding & Acc & 55.17 & \textbf{59.48} & 51.72 & 25.86 & 25.00 \\
~ & Professional Filming and Editing Techniques & Acc & \textbf{61.35} & 50.92 & 55.21 & 30.67 & 24.54 \\
\hline
~ & Overall & Acc & 64.33 & \textbf{64.58} & 54.37 & 33.94 & 28.01 \\
\bottomrule[1.2pt]
\end{tabular}
}
\caption{
Evaluation results on the VUE-CHAR benchmark, reporting temporal intersection-over-union (tIoU), spatial intersection-over-union (sIoU) and ASR errors in the \textit{Character} (Char.) and question accuracy in the \textit{Reasoning} (Rea.) track.}
\label{tab:plot_results_t}
\end{table}

\paragraph{Performance Evaluation.} \revise{
We present the detailed results in terms of grounding precision and reasoning accuracy in Table \ref{tab:plot_results_t}. Notably, \methodnew{}-Think significantly outperforms Gemini 3 Pro Preview \cite{comanici2025gemini} for spatial IoU on VUE-CHAR ($>42\%$) and achieves the lowest word error rate. For detailed analysis on the \textit{character} track, although Gemini 3 Pro Preview maintains a slight edge in overall accuracy due to strong base perception, \methodnew{}-Think excels on \textit{Speech, Audio, and Sound Effect Reasoning} ($74.43\%$ vs $66.03\%$) and \textit{Professional Filming and Editing Techniques} ($61.35\%$ vs $50.92\%$). We can conclude that the superior audio-visual alignment of \methodnew{}-Think ensures a better experience for understanding complex cinematic narratives and subtle character interactions. We show Exemplar cases of \methodnew{}-Think in Figure~\ref{fig:char} and Figure~\ref{fig:rea}. \methodnew{}-Think is able to densely localize the active speakers and recognize what they are talking about (see Figure~\ref{fig:char}).
\methodnew{}-Think also provides step-by-step reasoning to arrive at the correct answer, explicitly decomposing the complex plot understanding process into a sequence of interpretable intermediate inferences that mirror human-like temporal and causal reasoning over video content (see Figure~\ref{fig:rea}).
}

\begin{figure*}[!htbp]
    \centering
    \includegraphics[width=\linewidth]{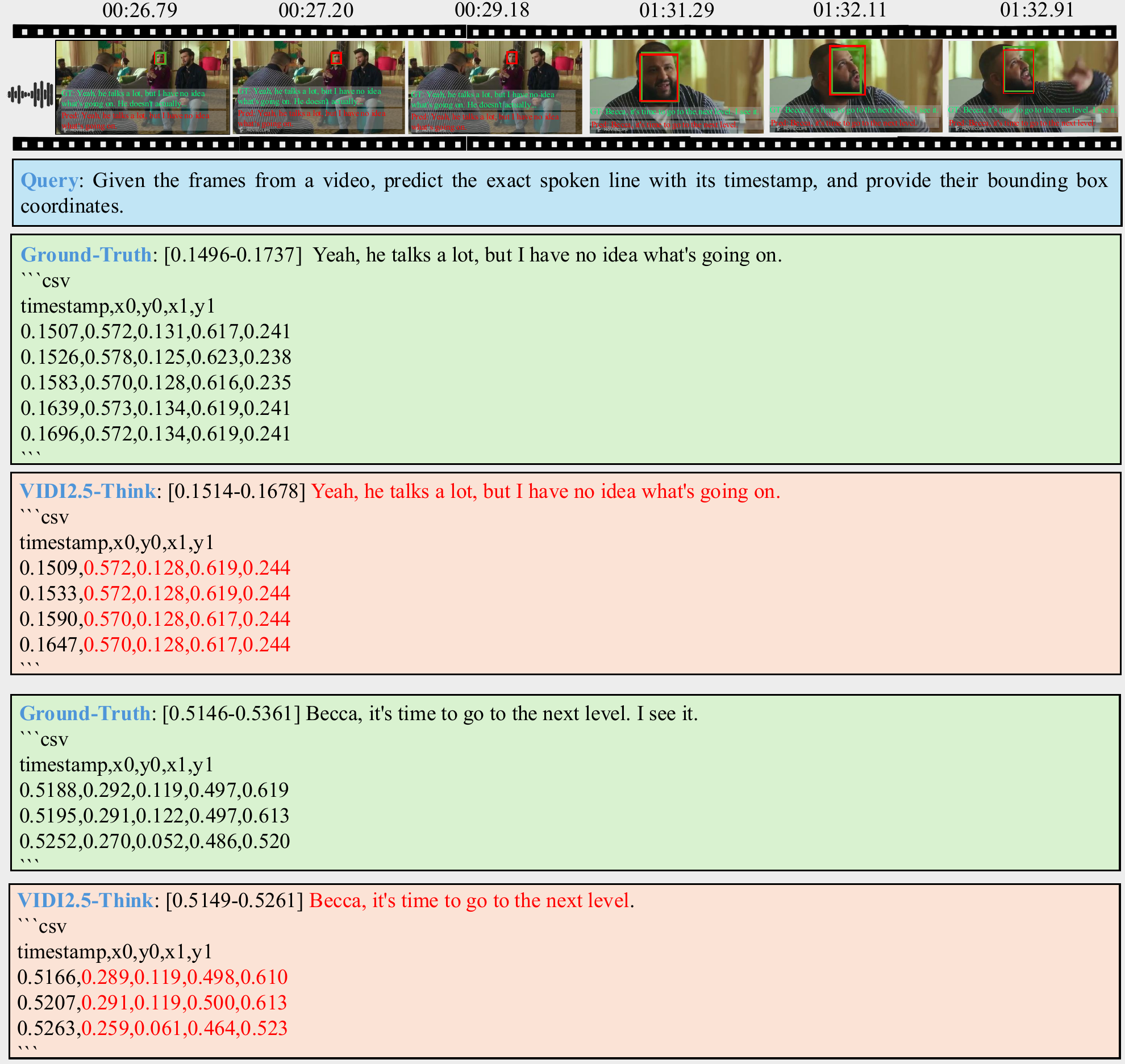}
    \caption{Exemplar cases of \methodnew{}-Think for dense speaker localization and speech recognition in videos. The ground-truth annotation and the model prediction are are marked with green and red colors in the input video for better visualization.}
    \label{fig:char}
\end{figure*}

\begin{figure*}[!htbp]
    \centering
    \includegraphics[width=\linewidth]{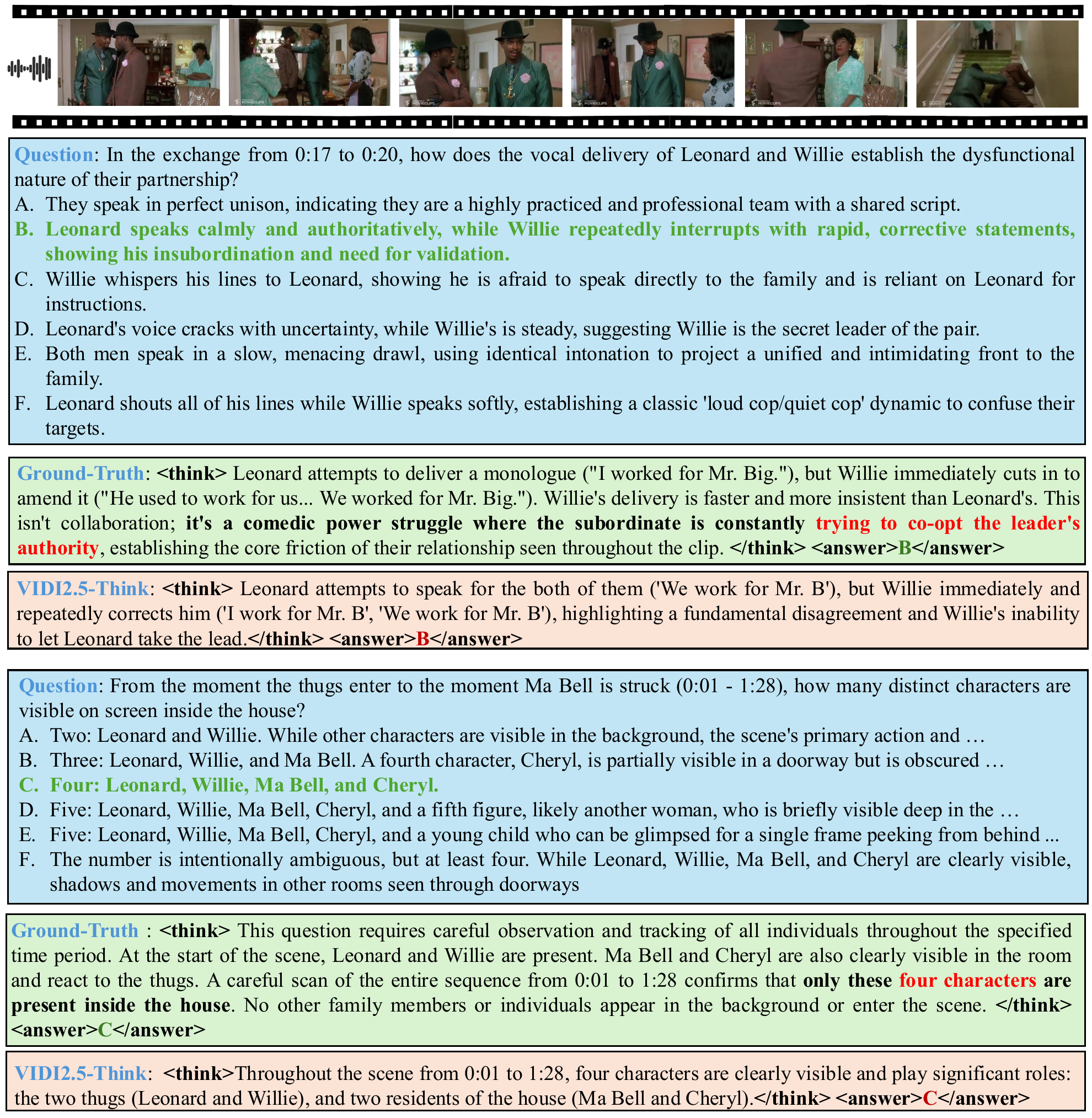}
    \caption{Exemplar cases of \methodnew{}-Think for plot reasoning in videos. \methodnew{}-Think provides step-by-step reasoning to arrive at the correct answer. The correct option is marked with green color for better visualization.}
    \label{fig:rea}
\end{figure*}
{
\setlength{\parindent}{0pt}
\setlength{\parskip}{0.4em}

\section{Applications}
\label{sec:applications}

Beyond standard multimodal understanding benchmarks, \methodnew{} can be adapted through post-training to support a broad range of downstream applications that require structured reasoning, long-horizon planning, and coordinated multimodal decision-making.
In this work, we present \textbf{video editing planning} as a representative case study to demonstrate these capabilities in a concrete and practical scenario.

\subsection{Video Editing Planning}

Video editing planning aims to determine a set of interdependent high-level editing decisions that collectively define how a final video is created from raw assets. Given a collection of raw assets (\eg, images and videos) and optional user prompts (\ie, user-specified editing requirements), the objective is to generate an editing plan that specifies the narrative structure, narration content, audio attributes, and visual editing intent.

In our setting, the editing plan is represented as a high-level textual or structured representation that unifies narrative structure, audio design, and visual intent. This representation enables downstream systems to perform audio synthesis, visual effect parameterization, and video rendering in a coordinated manner.

\subsection{Editing Plan Specification}

The editing plan serves as a semantic blueprint, defining a set of high-level decisions that together specify the structure, presentation, and intent of the final video. As shown in Figure~\ref{fig:application_planner}, these decisions act as a unified directive for downstream execution systems.

\paragraph{Narrative structure} defines the hierarchical and temporal orchestration of the video, encompassing clip selection, segment extraction, and segment ordering. It establishes the temporal and semantic backbone of the edit and governs how visual content is arranged to convey the intended narrative.

\paragraph{Voiceover content} specifies the narrated material aligned with the narrative structure, including semantic content, delivery style, and temporal alignment. It determines how narration supports the storyline and provides guidance for downstream speech synthesis.

\paragraph{Audio attributes} characterize auditory elements such as music style, mood, rhythmic profile, and speaker attributes, which contextualize and complement the visual narrative. This attribute set establishes the auditory framing of the video and promotes coherence between sound design and visual progression.

\paragraph{Visual editing intent} articulates stylistic and rhythmic presentation directives in semantic terms, including transition design, emphasis cues, and stylistic directives. This specification communicates how visual effects should be applied to support narrative structure and pacing.

Together, these decision categories define the specification space of the editing plan, capturing the intended structure and presentation of the video while remaining agnostic to execution-specific details.

\begin{figure*}[!htbp]
    \vspace{-3em}
    \centering
    \setlength{\abovecaptionskip}{-10pt}
    \includegraphics[height=0.95\textheight, keepaspectratio]{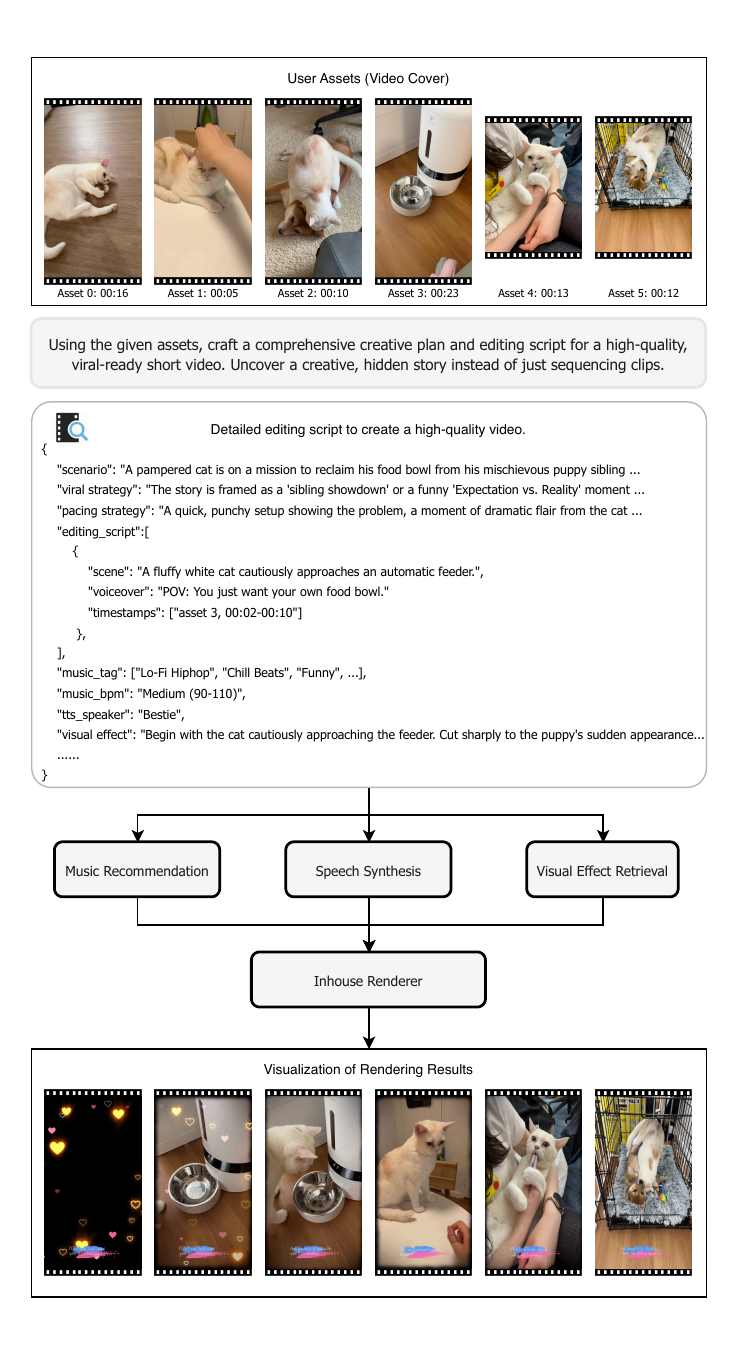}
    \caption{Overview of the \vidiedit{} execution pipeline.
The generated editing plan specifies high-level semantic intent, including narrative structure, voiceover content, audio attributes, and visual editing directives, which is subsequently translated into concrete operations, such as music recommendation, speech synthesis, visual effect retrieval, and video rendering. This design explicitly separates high-level planning from execution, allowing the model to focus on semantic reasoning while downstream systems handle implementation.}
    \label{fig:application_planner}
\end{figure*}

\subsection{Post-Training for Editing Plan Generation}
To support video editing planning, we post-train \methodnew{} to align its multimodal reasoning capabilities with the structured requirements of editing plan generation. We refer to this post-trained configuration as \vidiedit{}. This post-training stage does not modify the underlying model architecture; instead, it adapts the pre-trained representations to produce editing plans that conform to the specification described above.

Post-training leverages additional task-specific supervision in the form of multimodal inputs paired with textual or structured editing plans. The supervision signal emphasizes adherence to the planning format and semantic consistency across narrative structure, voiceover content, audio attributes, and visual editing intent, enabling the model to express its reasoning outcomes in a coherent planning-oriented representation.

\subsection{Editing Plan Implementation}

The editing plans generated by \vidiedit{} are realized through a structured execution pipeline that translates high-level planning intent into concrete editing actions. As shown in Figure~\ref{fig:application_planner}, the generated plans, expressed in natural language or structured text, are first processed by a translation stage that maps semantic specifications to executable components. 

During this interpretation stage, different aspects of the editing plan are grounded through specialized modules. Music attributes specified in the plan are retrieved from a curated music database, while the narration content is synthesized by a text-to-speech (TTS) generator according to the speaker attributes. The visual editing intent is mapped to the corresponding visual effect primitives through a dedicated effect retrieval and parameterization module.

The resulting audio assets and visual effect parameters are then passed to an in-house rendering system, which integrates video clips, synthesized audio, textual overlays, and visual effects to produce the final high-quality videos. This modular execution pipeline cleanly separates high-level editing decisions from their realization, allowing the planning model to focus on semantic reasoning while execution components handle asset selection, synthesis, and rendering.

Under this design, the planning-and-execution pipeline can be further extended into an editing agent architecture, where the \vidiedit{} serves as a high-level decision-making module and the execution pipeline grounds these decisions into concrete editing actions.

\subsection{Qualitative Examples}
We present qualitative examples to examine the editing plans produced by \vidiedit{} and to illustrate how high-level editing intent is reflected in the resulting outputs. Figures~\ref{fig:application_planner_zoo}~\ref{fig:application_planner_tokyo}~\ref{fig:application_planner_seattle}~\ref{fig:application_planner_mix} show representative rendered video frames from which narrative structure, clip arrangement, and voiceover progression can be inferred.

In these examples, \vidiedit{} generates coherent editing plans that organize raw video clips into an explicit temporal structure. The clips are arranged according to the planned storyline, including context setup (\eg, an opening hook), content development, and a concluding wrap-up. Voiceover content is aligned with this temporal structure, providing a consistent narrative thread that follows the planned progression of the video.

While the figures visualize clip ordering and voiceover progression, other planning attributes, such as audio design and visual effect specifications, are not directly observable in static frames. To complement these visualizations, additional qualitative results and example videos are provided on the project website, along with an interactive interface for exploring the model’s capabilities.

\begin{figure*}[!htbp]
    \centering
    \setlength{\abovecaptionskip}{-10pt}
    \includegraphics[height=0.9\textheight, keepaspectratio]{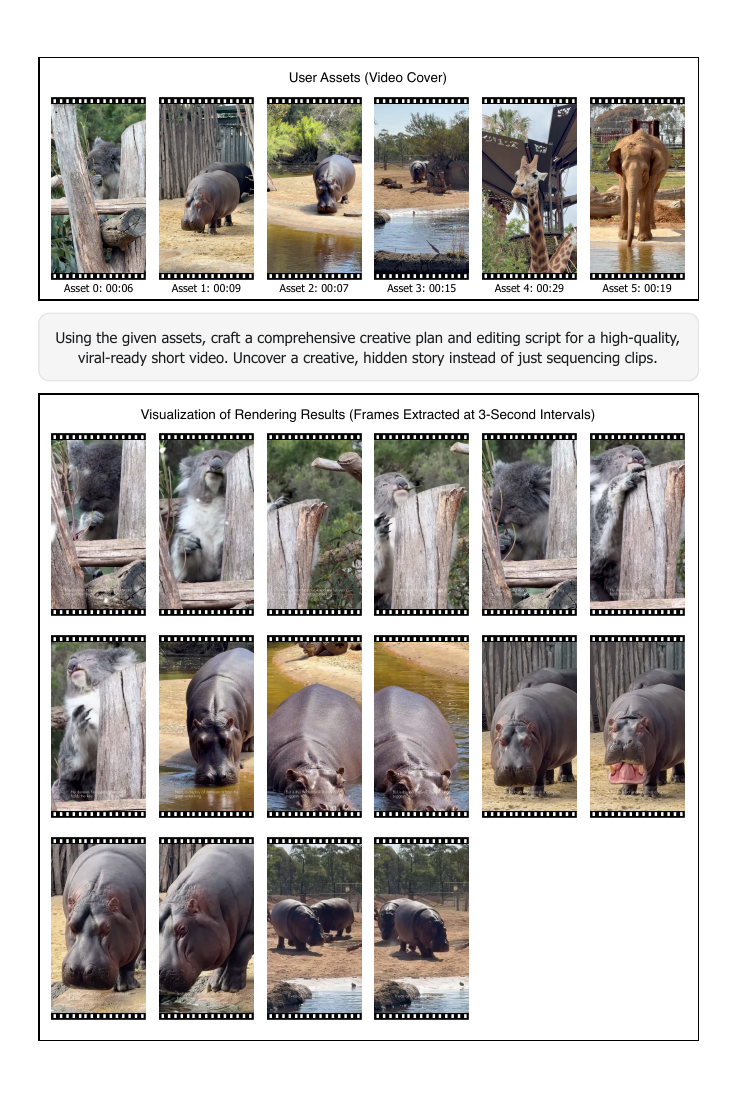}
    \caption{Qualitative example of an editing plan generated by \vidiedit{} for a single-topic scenario (zoo animals). All input assets are captured within the same zoo environment. The model identifies salient moments across multiple animal clips and arranges them into a coherent narrative with implicit temporal structure, rather than merely concatenating visually similar footage.}
    \label{fig:application_planner_zoo}
\end{figure*}

\begin{figure*}[!htbp]
    \centering
    \setlength{\abovecaptionskip}{-10pt}
    \includegraphics[height=0.9\textheight, keepaspectratio]{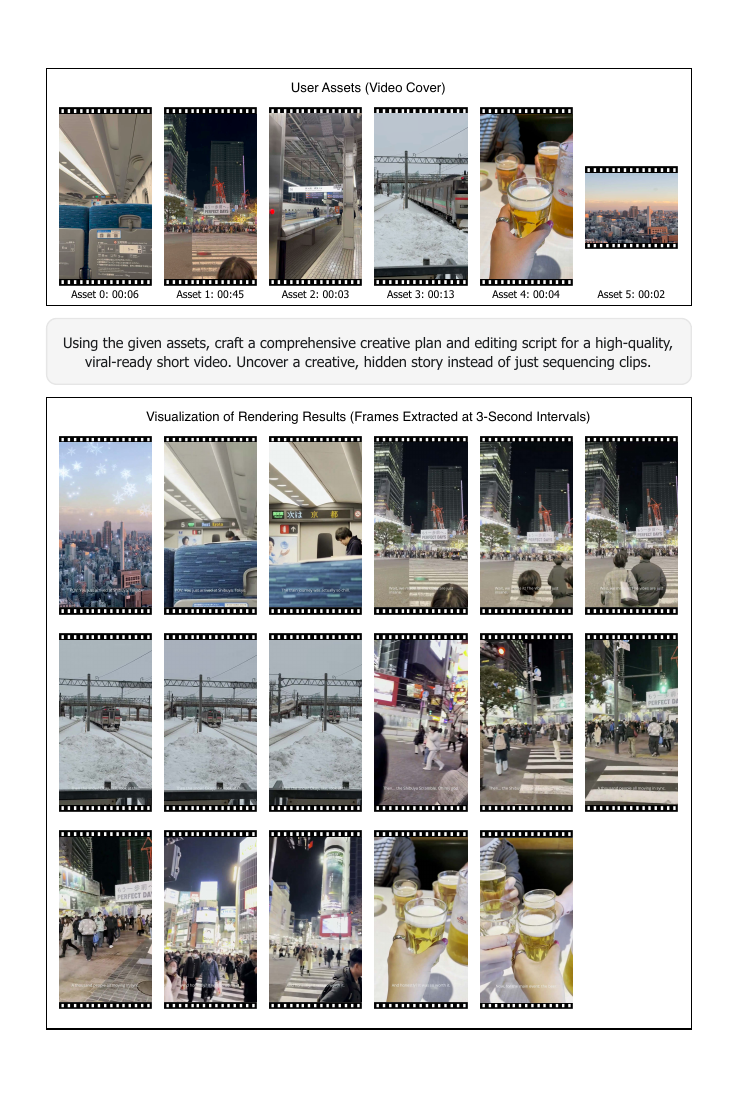}
    \caption{Qualitative example of an editing plan generated by \vidiedit{} for a single-topic urban travel scenario. Given heterogeneous user assets from a city visit (\eg, transportation, streets, nightlife, and social moments), the model constructs a unified storyline that captures temporal progression and experiential flow across diverse scenes.}
    \label{fig:application_planner_tokyo}
\end{figure*}

\begin{figure*}[!htbp]
    \centering
    \setlength{\abovecaptionskip}{-10pt}
    \includegraphics[height=0.9\textheight, keepaspectratio]{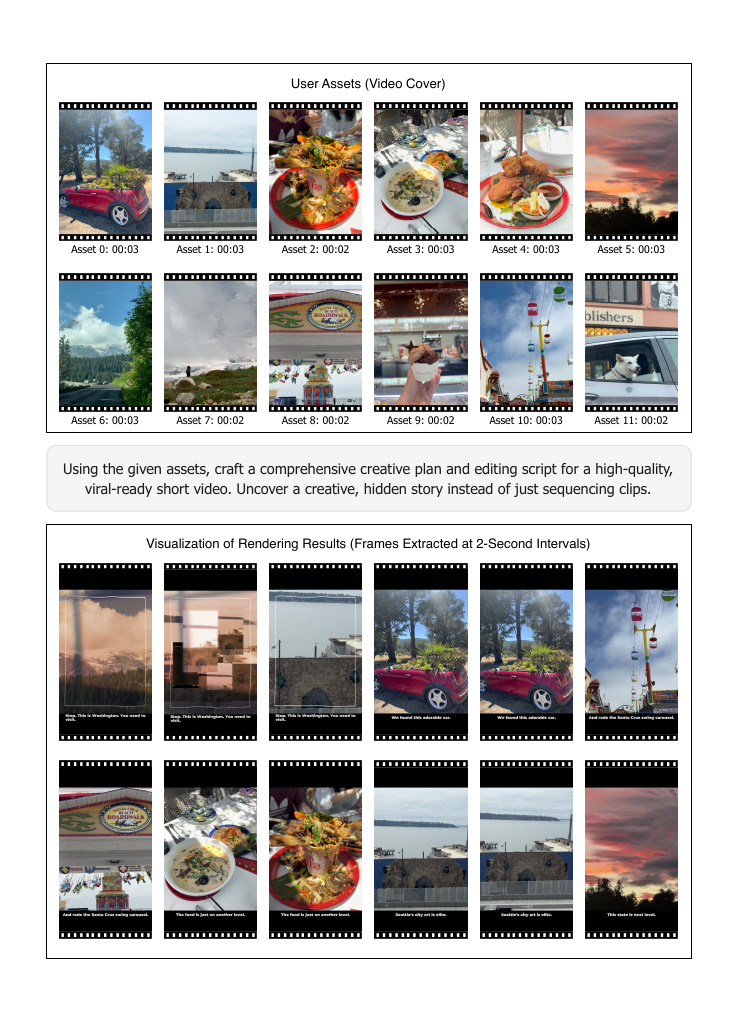}
    \caption{Qualitative example of an editing plan generated by \vidiedit{} for a single-topic urban scenario set in Seattle. Each input consists of only a few seconds, providing limited temporal context in isolation. Despite this constraint, the model infers high-level semantic relationships across clips and organizes them into a coherent narrative, demonstrating robustness to sparse and fragmented inputs.}
    \label{fig:application_planner_seattle}
\end{figure*}

\begin{figure*}[!htbp]
    \centering
    \setlength{\abovecaptionskip}{-10pt}
    \includegraphics[height=0.9\textheight, keepaspectratio]{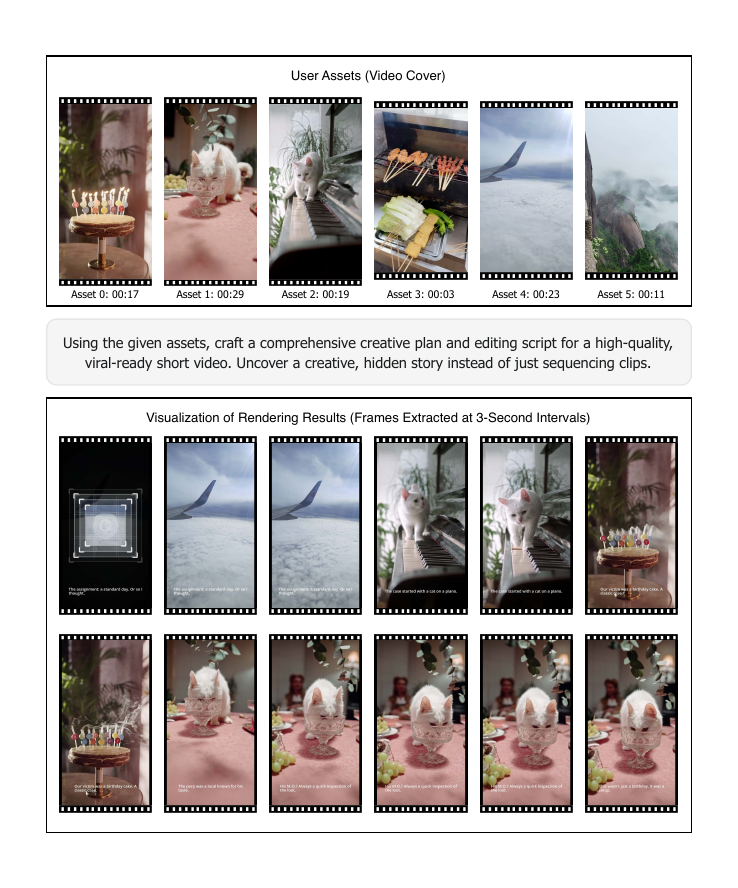}
    \caption{Qualitative example of an editing plan generated by \vidiedit{} for a mixed-content scenario. The input assets are semantically diverse and lack an obvious temporal or thematic order. \vidiedit{} selectively identifies and arranges appropriate clips to construct a coherent storyline, demonstrating its ability to perform high-level narrative planning beyond low-level visual similarity or naive clip concatenation.}
    \label{fig:application_planner_mix}
\end{figure*}

}
\clearpage
\section{Conclusion}
\label{sec:conclusion}
In this report, we first present \method{}, which achieves unprecedented end-to-end spatio-temporal grounding performance while maintaining strong video question answering capabilities.
It serves as a foundation model for next-generation video creation and understanding scenarios.
To evaluate models in realistic video editing contexts, we introduce VUE-STG, a comprehensive benchmark that advances prior spatio-temporal grounding datasets in four aspects: video duration, query format, annotation quality, and evaluation metrics.
We also upgrade our previous temporal retrieval benchmark to VUE-TR-V2, featuring a more balanced video-length distribution (including more long and ultra-long videos) and more realistic, user-style queries.
In addition, we evaluate \method{} on public video QA datasets to assess its general multimodal reasoning ability.
Notably, \method{} significantly outperforms proprietary systems such as Gemini 3 Pro Preview and GPT-5 on both VUE-STG and VUE-TR-V2, demonstrating state-of-the-art spatio-temporal understanding.
It also performs on par with the leading open-source video LLM of similar scale, \ie, Qwen2.5-VL-7B, on video QA.
Finally, we outline three promising real-world applications that \method{} could enable in future demonstrations.
Together, these results highlight the model's strong spatio-temporal understanding and multimodal reasoning capabilities, forming a robust foundation for intelligent, human-like video editing and creation systems.\\

\noindent In addition, we introduce the latest \methodnew{} that adopts RL training to consistently improve the performance of TR, STG and video QA over the \method{} base model. We also introduce a thinking model, \methodnew{}-Think, with inference-time scaling to handle complex reasoning scenarios in video creation. Specifically, we focus on plot understanding in this report and propose the VUE-PLOT benchmark to comprehensively evaluate the fine-grained perception and complex reasoning capability in this task. \methodnew{}-Think outperforms Gemini 3 Pro Preview on fine-grained character understanding by a large margin, and achieves comparable performance on complex plot reasoning tasks.\\

\noindent Beyond benchmark-driven evaluation, we further demonstrate that \methodnew{} can be adapted through post-training to support practical video creation workflows. Using video editing planning as a representative application, the post-trained variant \vidiedit{} generates structured editing plans that specify narrative structure, audio attributes, and visual editing intent from raw video assets. This application illustrates how the spatio-temporal understanding and multimodal reasoning capabilities of \methodnew{} can be translated into executable planning representations for real-world video editing systems.

\clearpage
\section{Contributors}
\label{sec:contributor}
\textbf{Core Contributors - Research - Foundation (alphabetical order)}\\
Chia-Wen Kuo, Dawei Du, Fan Chen, Guang Chen, Haoji Zhang, Sijie Zhu, Xin Gu, Zhenfang Chen. \\

\noindent \textbf{Core Contributors - Research - Application (alphabetical order)}\\
Feng Gao, Lu Xu, Wen Zhong, Ye Yuan, Yiming Cui. \\

\noindent \textbf{Core Contributors - Infrastructure (alphabetical order)} \\
Fanding Lei, Haojun Zhao, Tong Jin. \\

\noindent \textbf{Team Leads} \\
Longyin Wen\footnotemark[1], Xiaohui Shen. \\

\noindent \textbf{Contributors (alphabetical order)}\\
Chuang Huang, Jin Liu, Jingjing Zhuge, Lili Fang, Lingxi Zhang, Lu Guo, Lusha Li, Qihang Fan, Rachel Deng, Shaobo Fang, Shu Zhang, Stuart Siew, Weiyan Tao, Yicheng He, Zhihua Wu, Zuhua Lin.

\footnotetext[1]{Corresponding author.}
\renewcommand{\thefootnote}{\arabic{footnote}}

{
\small
\bibliographystyle{plain}
\bibliography{main}
}
\end{document}